\documentclass[letterpaper, 10 pt, conference]{ieeeconf}  

\IEEEoverridecommandlockouts                              

\pdfoutput=1

\usepackage{graphicx} 
\usepackage{amsmath} 
\usepackage{amssymb}  

\usepackage{amsthm}

\usepackage{bm}
\usepackage[dvipsnames]{color}
\usepackage{subfig}
\usepackage{psfrag}
\usepackage[linesnumbered,ruled,vlined]{algorithm2e}
\usepackage{array}
\usepackage[font=footnotesize]{caption}
\usepackage{xspace}
\usepackage{blindtext}
\usepackage{booktabs}
\graphicspath{{figures/}}

\title{\LARGE \bf
	Topological Information-Theoretic Belief Space Planning with Optimality Guarantees*
}

\author{Andrej Kitanov and Vadim Indelman
	\thanks{*This work was partially supported by the Israel Science Foundation.}
	\thanks{The authors are with the Department of Aerospace Engineering, Technion - Israel Institute of Technology, Haifa 32000, Israel. {\tt\small \{andrej, vadim.indelman\}@technion.ac.il}.}
}

\DeclareMathOperator*{\argmin}{arg\,min}
\DeclareMathOperator*{\argmax}{arg\,max}

\newcommand{\maxim}[2]{\ensuremath{\underset{#2}{#1}}}

\newcommand{\nrmsq}[2]{\ensuremath{{\parallel}{#2}{\parallel^2_{#1}}}}

\newcommand{\prob}[1]{\ensuremath{\mathbb{P}({#1})}}

\newcommand{\tmop}[1]{\ensuremath{\operatorname{#1}}}

\newcommand{\pruning}[1]
{\if\relax\detokenize{#1}\relax
  \ensuremath{\sigma_{prune}}%
\else
  \ensuremath{\mathbf{\sigma_{prune}=#1}}%
\fi
}
\newcommand{\merging}[1]
{\if\relax\detokenize{#1}\relax
  \ensuremath{\sigma_{merge}}%
\else
  \ensuremath{\mathbf{\sigma_{merge}=#1}}%
\fi
}

\newcommand{\tbsp}{\ensuremath{\texttt{t-BSP}}\xspace}

\definecolor{rowcolor1}{rgb}{1,1,1}
\definecolor{rowcolor2}{rgb}{0.8,0.8,0.8}

\newtheorem{theorem}{Theorem}

\theoremstyle{definition}
\newtheorem{definition}{Definition}
\theoremstyle{remark}

\AtBeginDocument{%
	\setlength{\textfloatsep}{5pt}
	\setlength{\floatsep}{5pt}
	\setlength{\intextsep}{5pt}
	\setlength{\belowdisplayskip}{5pt} \setlength{\belowdisplayshortskip}{5pt}
	\setlength{\abovedisplayskip}{5pt} \setlength{\abovedisplayshortskip}{5pt}
}

\SetKwInput{KwData}{Parameters}

\begin{document}

\maketitle
\thispagestyle{empty}
\pagestyle{empty}

\begin{abstract}

Determining a globally optimal solution of belief space planning (BSP) in high-dimensional state spaces is computationally expensive, as it involves belief propagation and objective function evaluation for each candidate action. Our recently introduced  topological belief space planning (\tbsp) \cite{Kitanov18icra} instead  performs decision making considering only topologies of factor graphs that correspond to posterior future beliefs.
In this paper we contribute to this body of work a novel method for efficiently determining error bounds of \tbsp, thereby providing global optimality guarantees or uncertainty margin of its solution.  The bounds are given with respect to an optimal solution of information theoretic BSP considering the previously introduced topological metric which is based on the number of spanning trees. In realistic and synthetic simulations, we analyze tightness of these bounds and show empirically how this metric is closely related to another computationally more efficient \tbsp metric, an approximation of the von Neumann entropy of a graph, which can achieve online performance.

\end{abstract}

\section{INTRODUCTION}

A core property of an intelligent autonomous robotic system is its ability to autonomously make smart online decisions under uncertainty. The corresponding problem is known as belief space planning (BSP) and can be seen as a joint control and estimation problem in which an agent (robot) has to find optimal control according to a specific task-related objective, which itself has to be estimated while accounting for different sources of uncertainty, e.g.~due to stochastic sensing, motion or environment.  
Some interesting instantiations of this problem are active SLAM (e.g.~\cite{Stachniss04iros,Huang05icra,Du11aim,Valencia12iros,Kim14ijrr,Indelman15ijrr,Atanasov15icra, Kopitkov17ijrr, Regev17arj, Indelman15ijrr}), active perception \cite{Bajcsy88ieee}, sensor deployment  and measurement selection \cite{Joshi09tsp}\cite{Bian06ipsn}\cite{Hovland97icra}, graph reduction and prunning \cite{Kretzschmar12ijrr, CarlevarisBianco14tro}. 

The general BSP problem can be naturally represented by a Partially-Observable  Markov  Decision  Process (POMDP). 
Determining globally optimal solutions to the POMDP problem is computationally intractable for most but simplest real-world problems mainly due to high dimensionality of estimated states. Computational complexity remains an issue even with simplifying assumptions on the probability distribution of the states, finite planning horizons, discrete  states, actions or observations  \cite{Papadimitriou87math}. As a result, a large subset of prior work has focused on approximately  solving  the  POMDP  problem to provide better scalability. Some examples include using  sampling-based  motion  planners in  Gaussian  belief spaces (e.g.~\cite{Kavraki96tra, Prentice09ijrr, AghaMohammadi14ijrr}), local optimization methods for continuous state spaces (e.g.~\cite{Indelman15ijrr, Platt10rss, VanDenBerg12ijrr}) or point-based value iteration (e.g.~\cite{Pineau06jair, Porta06jmlr,Kurniawati08rss}). 

An information-theoretic cost in  an  objective  function of BSP is some measure of system uncertainty (information), typically (conditional) entropy or mutual information. Determining it requires evaluating  the  expected  posterior  belief upon  action  execution. For Gaussian  distributions  the  corresponding  computations usually involve  calculating  a  determinant  of a posteriori covariance  (information) matrix whose complexity is $O(n^3)$ in general case, where $n$ is a state dimension. Moreover,  these  calculations need to be performed  for each candidate  action. In \cite{Indelman15ijrr} this challenge was addressed by resorting to information form and utilizing sparsity; however, calculations still involve expensive access to marginal probability distributions. The rAMDL approach \cite{Kopitkov17ijrr} performs a one-time calculation of marginal covariance recovery of variables involved in the candidate actions, and then applies an augmented matrix determinant lemma (AMDL) to efficiently evaluate the information-theoretic cost for each candidate action. Nevertheless, that approach still requires recovery of appropriate marginal covariances, the complexity of which depends on state-dimensionality and system sparsity. 

Notice that optimality of the above solutions is only achieved under the  assumption of perfectly known system's probabilistic models, (e.g. predicting future observations, data associations, noise distributions etc.) 
which is rarely the case in unknown environments. So, by an appropriate approximation of the problem, the loss of accuracy in BSP can be even in the range of modelling errors, but speed much higher. More importantly, if we can quantify the error of making such approximation, optimality guarantees can be established. If needed, the optimal solution can still be  obtained, but this time, by evaluating only a subset of  candidate actions, while discarding the rest, as proposed in \cite{Elimelech17isrr}. Doing so will generally reduce the number of variables for which the marginal covariance needs to be recovered.

The second important observation is that in order to choose an optimal action, exact value of the objective function is not required as long as the actions can be sorted from best to worst. In particular, changing these values (due to a simplified problem representation) without changing the order of actions does not change the action selection. This concept is called \textit{action consistency} and was recently introduced in \cite{Elimelech17isrr, Elimelech17icra, Elimelech17iros}, building upon \cite{Indelman16ral, Indelman15acc}. These works apply this concept to find a simplified representation of the belief by doing sparsification that keeps decision making action consistent, or with a bounded error but assume myopic actions and one-row action Jacobians.
In the general case, however, a simplified representation  may lead to some sacrifice in performance. 



Our work is motivated by the results  of \cite{Khosoussi18ijrr} in the context of measurement  selection  and  pose-graph  pruning  problems  in  SLAM that   characterizes  the  impact  of  the  graph topology  of SLAM (described by weighted tree-connectivity metric) on  the  estimation  reliability.
We extended these results to BSP problem in \cite{Kitanov18icra} and introduced a \emph{topological BSP} (\tbsp) concept and new topological metrics $s(\mathcal{U})$ to approximate the solution to BSP. One \tbsp metric we introduced is based on the number of spanning trees of the topological posterior factor graph representation and the other on its von Neumann entropy. We showed they both have strong correlation with the information-theoretic cost and lead to fast convergence of the BSP solution when realized in anytime algorithmic manner. However, global optimality guarantees were not established so we could not say when optimal solution was actually reached. 
This work provides theoretical foundations of \tbsp and improvements of its time performance. Its main contributions are: 
\begin{enumerate}
	\item We derive bounds of the information-theoretic cost in BSP which can be calculated online, i.e. with a small additional cost to the topological metric, and used to provide global optimality guarantees or uncertainty margin of \tbsp solution.
	\item We extend the definition of action  consistent state  approximations proposed in \cite{Elimelech17isrr} to \tbsp and action consistent objective approximations. More, we address non-myopic actions and vector measurements.
	\item In realistic simulations of active pose SLAM and random synthetic pose graphs, we analyze time complexity and action consistency of \tbsp.
\end{enumerate}


\section{Notations and Problem Formulation}

Decision making under uncertainty can be formulated as a solution to BSP or stochastic control problem where optimal non-myopic control action $\mathcal{U}^{\star}_{k:k+L-1}$ at planning time $k$ is found with respect to the 
objective function $J$ related to the design task as 
\begin{align}
 & \mathcal{U}^{\star}_{k:k+L-1}  =  \textstyle \argmin_{\mathcal{U}} J(\mathcal{U}),\text{ where for } \mathcal{U} = \mathcal{U}_{k : k + L - 1} \label{eq:OptimalControl}  \nonumber \\
& J (\mathcal{U})= \underset{\ \mathcal{Z}}{\mathbb{E}}\left\lbrace \sum_{l = 0}^{L - 1} c_l \left[ b (X_{k + l}),
\mathcal{U}_{k + l}\right]  + c_L \left[ b (X_{k + L})\right] \right\rbrace.
\end{align}
As can be seen, the BSP problem is an instance of POMDP because the state of the system $X$ is not directly observable by the controller, but through a set of stochastic measurements $\mathcal{Z}$ from which the future posterior beliefs $b (X_{k + l})$ must be inferred upon optimization. Expectation in \eqref{eq:OptimalControl} though is taken with respect to future (unknown) observations $\mathcal{Z}_{k+1:k+L}$. The optimal solution of BSP provides a control strategy $\mathcal{U}^{\star}_{k:k+L-1}$ for $L$ look-ahead steps, but $L$ can generally vary among control actions.
We shall sometimes omit the explicit notation and use $\mathcal{U}^{\star} = \mathcal{U}^{\star}_{k:k+L-1}$ to denote the optimal control policy in a given planning session (at time $k$).
The objective function in its general form reflects the design task through immediate cost functions $c_{l}$, which depend on the belief evolution $b[X_{k+l}]$ (to be defined) and a control action $\mathcal{U}_{k+l}$ applied at time $t_{k+l}$, and through a final  cost $c_{L}$. For example, the cost functions can be chosen to minimize trajectory uncertainty, time or energy required to reach a goal, state uncertainty of variables of interest at some specific time instant etc. In information-theoretic BSP, one is interested in state uncertainty minimization which can be expressed through some information-theoretic cost $c(.)$ (see e.g. \cite{Carrillo12icra}). This type of cost functions is usually computationally the most expensive to optimize in many BSP problems in robotics.
Among them, active SLAM is considered as one of the most general BSP problems because it includes simultaneously estimating the robot's pose and the map of the environment, while planning for the path that improves both estimates.

In this work we consider an active pose SLAM problem and minimizing path uncertainty (quantified by joint entropy) to study the main aspects of \tbsp. Extension to a feature-based SLAM and map entropy is possible under the proposed framework as long as the measurements can be expressed in the form of binary relations between states. This is quite common since robot's landmark measurements are often given relative to the robot's pose, e.g. as range or bearing measurements.
For clarity we only consider the final state cost term $\mathbb{E}\left[  c_L(b[X_{k+L}])\right]$, i.e. the joint entropy at the end of planning horizon. The immediate information-theoretic cost functions can be treated in a similar way.
Considering the belief is modeled by a multivariate Gaussian distribution and taking the common maximum likelihood observations assumption \cite{Platt10rss}, the objective function becomes 
\begin{align}
J(\mathcal{U})=N/2\ \text{ln} (2\pi \text{e}) + 1/2\ \text{ln}\lvert \Sigma(X_{k+L})\rvert,
\label{eq:ObjFunctionSep}
\end{align}
where $\Sigma(X_{k+L})$ denotes the estimated covariance of the robot's belief $b[X_{k+L}]$, and $N$ dimension of the state $X_{k+L}$. Notice that minimizing the global entropy \eqref{eq:ObjFunctionSep} corresponds to maximizing the total information gain obtained by executing this trajectory, i.e. maximizing the estimation accuracy.

Solving the optimization problem \eqref{eq:OptimalControl} explicitly would require belief propagation for a given control and evaluating an information-theoretic objective which we want to avoid and use a topological approach instead. Let us first look how the belief evolves over time by separating controls and observations to those obtained by  planning time $t_k$ and to future controls and observations after $L$ look-ahead steps. 

Let $\prob{X_{k}|\mathcal{H}_k}$ represent the posterior probability density function (pdf) at planning time $t_k$ over states of interest $X_{k}$ of the robot. In the pose SLAM framework states of interest are robot's current and past poses, i.e. $X_{k} = \{x_0, x_1, \ldots, x_k\}$. History $\mathcal{H}_k \doteq \{ \mathcal{Z}_{1:k}, \mathcal{U}_{0:k-1} \}$ contains all observations $\mathcal{Z}_{1:k}$ and controls $\mathcal{U}_{0:k-1}$ by time $t_k$. Consider conventional state transition and observation models 
\begin{equation}
x_{i+1} = f(x_i, u_i, w_i) \ , \ 
z_{i,j} = g(x_i,x_j,v_{i,j}), 
\label{eq:MotionObsModel}
\end{equation}
with zero-mean Gaussian process and measurement noise $w_i \sim N(0, \Omega_w^{-1})$ and $v_{i,j} \sim N(0, \Omega_{vij}^{-1})$, and with known information matrices $\Omega_w$ and $\Omega_{vij}$.
Belief $\prob{X_{k}|\mathcal{H}_k}$ is then 
\begin{equation}
\prob{X_{k}|\mathcal{H}_k} \!\propto\! \prob{x_0}\prod_{i=1}^{k} \prob{x_{i}|x_{i-1}, u_{i-1}}\prob{Z_{i}|X_{i}}.\label{eq:priorPDF}
\end{equation}
Let the future sampled states of the robot along one of its candidate paths $\mathcal{P}$ generated at planning time $t_k$ be $\{ x_{k+1},\ldots,x_{k+L}\}$.
The future posterior belief $b[X_{k+L}] = \prob{X_{k+L}|\mathcal{Z}_{1:k+L}, \mathcal{U}_{0:k+L-1}}$ that would be obtained by following the path $\mathcal{P}$,  can be  written in terms of the belief at planning time  $\prob{X_{k}|\mathcal{H}_k}$ and the corresponding state transition and observation models as 
%
\begin{equation}
b[X_{k+L}] = \prob{X_{k}|\mathcal{H}_k} \!\!\!\! \prod_{l=k+1}^{k+L(\mathcal{P})} \!\! \prob{x_{l}| x_{l-1}, u_{l-1}}  \prob{Z_{l} | X_{l}},
\label{eq:futureBelief}
\end{equation}
where $\mathcal{U}_{k:k+L-1}$ and $\mathcal{Z}_{k+1:k+L}$ represent controls and (unknown) observations, respectively, to be acquired by following the path $\mathcal{P}$. Note, to reduce notational clutter, we omit the explicit notation of the path $\mathcal{P}$ from (\ref{eq:futureBelief}).
%


The robot's belief (\ref{eq:futureBelief}) can be represented by a factor graph graphical model \cite{Kschischang01it} and assigned a topology represented by a simple undirected graph $G=(V,E)$ in the case of pose SLAM as we described in \cite{Kitanov18icra}, such that graph nodes $V$ represent robot's poses and edges $E$ pose constraints between them. We have studied BSP structural properties based on the eigenvalues $\{\hat{\lambda}_{i}\}_{i=1}^{n}$ of the normalized Laplacian matrix $\hat{L}$ associated with the graph $G$ and proposed a topological metric $s_{\tmop{VN}}(G)$, the Von Neumann entropy of $G$, and its simplification $\hat{s}_{\tmop{VN}}(G)$ by node degrees $d$, to approximate the solution of the original optimization problem \eqref{eq:OptimalControl} with the control action $\mathcal{U}$ that maximizes the metric \eqref{eq:VNsigs}.
\begin{eqnarray}
\displaystyle s_{\tmop{VN}}(G)  =  - \sum_{i=1}^{n} {\hat{\lambda}_{i}}/{2}
\ln ({\hat{\lambda}_{i}}/{2}),\\ 
\hat{s}_{\tmop{VN}}(G) \approx n/2 \ln 2 - 1/2 \sum_{( i,j ) \in E} {1}/{[	d ( i ) d ( j )]} \label{eq:VNsigs}
\end{eqnarray}
The second topological metric we considered in BSP is based on the number of spanning trees $t(G)$ of a graph $G$
\begin{equation}
s_{ST}(G) = {3}/{2} \ln t(G) + {n}/{2} [\text{ln}\ \lvert \Omega_{vij} \rvert -\text{ln}(2\pi e)^\kappa], \label{eq:STsig}
\end{equation}
normalized in such a way to account for different path lengths $n$ and dimension of the robot's poses, i.e. $\kappa = 3$ in 2D and $\kappa = 6$ in 3D active pose SLAM, that maintains a high correlation with the information-theoretic cost \eqref{eq:ObjFunctionSep}.

\section{Approach}

In this paper, we investigate error bounds of \tbsp on the active 2D pose SLAM problem. 
We first revise results on passive SLAM reliability regarding its graph structure described by a topological metric $\tau(G) \doteq \text{ln}(t(G))$ proposed in \cite{Khosoussi15} and then we show how these results can be extended to a BSP problem to provide online performance guarantees of \tbsp.

\subsection{Reliability of SLAM topology}

In the context of Maximum Likelihood Estimation (MLE) in pose SLAM, the optimal set of poses ${X_k}$ for which the belief \eqref{eq:priorPDF} is maximized can be obtained by fixing one of the poses, e.g. $x_0$, and treating the rest as unknown (or with an uninformative prior) while minimizing the sum of weighted squared errors between predicted and measured relative poses\footnote{Since the state $x_0$ is considered deterministic, here we estimate the rest of the variables $X_k = \{x_1, x_2, ..., x_k\}$ and $I(X_k)$ denotes their joint information matrix from now on.}, i.e. 
{\small
	\begin{equation}
	\hat{X}_k = \maxim{\argmax}{X_k}\ \prob{\Delta_k|X_k} \equiv \maxim {\argmin}{X_k} \nrmsq{\Sigma^{-1}}{\Delta_k - h(X_k)}.
	\end{equation}
}
In this formulation, a measurement $\Delta_k$ represents a vector of $m$ stacked relative pose measurements $z_{i,j}^r \in SE(2),\ r=1,2,\ldots,m$
from motion and observation model \eqref{eq:MotionObsModel} at time $t_k$ with $m = |E|$, the number of edges in the topological graph of the belief \eqref{eq:priorPDF}. 
Relative pose measurements in pose SLAM resulting from state transitions can be obtained by the motion composition
$z_{i + 1, i} (x_i, x_{i+1}) = \ominus x_i \oplus x_{i+1} = \ominus x_i \oplus f (x_i, u_i, w_i)$.
In this work, we assume independent relative pose measurements with additive noises
\begin{equation}
\Delta_k = h(X_k) + \nu_k, \ \nu_k \sim N(0, \Sigma^{-1}).
\end{equation}
For simplicity, we also assume a 2D pose SLAM setting in which all relative positions and orientations between poses $x_i$ and $x_j$ have equal variance, $\sigma_p^{2}$ and $\sigma_{\theta}^{2}$ respectively, i.e.~$\Omega_{\nu_{i,j}} = \text{diag}(\sigma_p^{-2}, \sigma_p^{-2}, \sigma_{\theta}^{-2})$. 
Measurement noise covariance $\Sigma$ in that case can be written as a diagonal matrix $\Sigma =  \text{diag}(\sigma_p^{2} I_{2m}, \sigma_\theta^{2} I_{m})$ by reordering elements of $\Delta_k$.
%

\begin{figure}[tbph]
\centering
\includegraphics[width=0.7\linewidth]{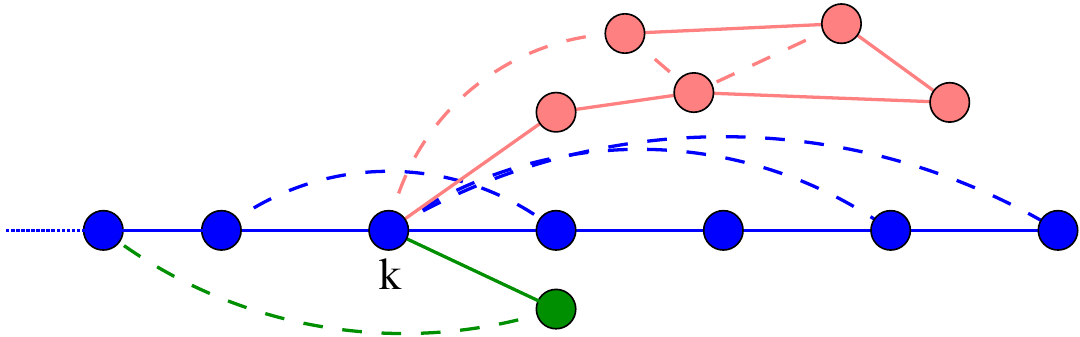}
\caption{BSP vs. measurement selection problem:	In measurement selection problems, the robot considers a single path and its factor graph (e.g. marked with blue color) at planning time $k$ and which subset of measurements (marked with blue dashed lines) to take. Pose samples $X_{k+L}$ are fixed and minimizing entropy \eqref{eq:ObjFunctionSep} by a measurements subset (action) is the same as maximizing $\text{ln} |I(X_{k+L})| - \text{ln} | I_o(X_{k+L}) |$. Its bounds as can be seen from \eqref{eq:detInfBounds} depend only on the topology and $\xi$, i.e.~all actions share the same pose variables $X_{k+L}$ which can be considered constant in optimization. 
However, in BSP where a robot needs to compare different paths $\mathcal{U}$ corresponding to different factor graphs (e.g. red and green), this is not true anymore. To determine the best action according to \eqref{eq:ObjFunctionSep}, one has to account additionally for different path geometries and path lengths and larger variety of topologies.} 
\label{fig:graph}
\end{figure}

The information matrix $I(X_k)$ of the MLE is
$I(X_k) = H^T \Sigma^{-1} H$ \cite{Sorenson1980},
where $H = \left.{\partial h}/{\partial X_k}\right.$ is a measurement Jacobian.
$I(X_k)$ evaluated at the true value of $X_k$ is known as Fisher Information Matrix (FIM) and its inverse the Cram\'{e}r-Rao lower bound (CRLB).
Commonly, FIM is approximated with $I(\hat{X}_k)$.
In \cite{Khosoussi15} bounds of the determinant of $I(X_k)$ are expressed in terms of pose SLAM graph topology, geometry and noise as
\begin{align}
3\tau(G)&+\ln \text{det} I_o(X_k) \leq \ln \text{det} I(X_k) \nonumber\\
&\leq 2 \tau(G) + \ln \text{det} (\tilde{L} + \Psi I) + \ln \text{det} I_o(X_k), \label{eq:detInfBounds}
\end{align}
where $\tilde{L}$ is a reduced Laplacian of a graph $G$ obtained by removing an  arbitrary $r$-th row and $r$-th column of the graph Laplacian $L$, $I_o(X_k)$ estimated information matrix based on the odometry measurements $\Delta_k = \{z_{i+1,i}, i < k\},$ and $\Psi \doteq \xi^2 \text{dist}_{max}^2$ where $\xi = \sigma_\theta/\sigma_p$, and $\text{dist}_{max}^2 = \max_{i \in V} \sum_{(i,j \in E)} \|x_i - x_j\|^2$. 
Notice that generally $\Psi$ depends on the noise variances, geometry and topology of the SLAM graph.
In \cite{Khosoussi15} these bounds are not used except to prove the limiting case, $\Psi \rightarrow 0.$ 
Therefore, for small values of $\Psi$ a good approximation of $\ln \text{det} I(X_k)$ is its bound that depends on $\tau(G)$, i.e. lower and upper bounds become tight. 
Even when it is not negligible, if there is only one path realization to consider as in passive SLAM, information gain of adding relative pose measurements with constant noise distribution to a SLAM odometry graph is solely characterized by graph $G$ topology, i.e. $\Psi = \Psi(G)$. Similar logic applies to graph prunning and measurement selection problem \cite{Khosoussi18ijrr} where all graphs are only subgraphs of the original graph with the same embedding in metric space.
To demonstrate why we cannot use the same metric $\tau(G)$ in BSP problems nor guarantee optimality using the bounds on $I(X_k)$, consider the example given in Fig. \ref{fig:graph}. 

So, in BSP we need to consider \textit{different} path realizations $X_{k+L}$ and therefore $\Psi = \Psi(\mathcal{U}) = \Psi(X_{k+L}(\mathcal{U}), G(\mathcal{U}))$, in contrast to graph pruning and measurement selection. 
Notice that we do not need to propagate belief in planning for determining $\Psi$ under  ML observations assumption, i.e. $\mathbb{E}[X_{k+L}] = [\hat{X}_k\ x_{k+1} \ldots x_{k+L}]$ where future sampled poses from the path corresponding to action $\mathcal{U}$ are added to the prior state estimate $\hat{X}_k$ which is the same for all actions.  

\subsection{Decision making via \tbsp}
The best control action obtained by solving the decision making problem using either of the topological metrics $s\in \{s_{ST}, s_{VN}, \hat{s}_{VN}\}$ is given by 
\begin{equation}
\mathcal{\hat{U}} = \argmax_{\mathcal{U}} s(\mathcal{U}).\label{eq:selectedCtrl}
\end{equation}
While the above topological metrics exhibit strong correlation with the information theoretic cost \cite{Kitanov18icra}, in the general case, the obtained best action $\mathcal{\hat{U}}$ may be somewhat different than the optimal action $\mathcal{U^{\star}}$ from (\ref{eq:OptimalControl}), leading to some error in the quality of solution.


We adopt the definition of action  consistent state  approximations proposed in \cite{Elimelech17isrr} and modify it to support \tbsp and action consistent objective approximation in the following way.
\begin{definition}
	The error of \tbsp is
	\begin{equation}
	\epsilon(J, s) \doteq \vert  J(\mathcal{U}^{\star}) - f[s (\mathcal{\hat{U}})] \vert, \label{eq:error}
	\end{equation}
	where $\mathcal{\hat{U}} = \argmax_{\mathcal{U}} s(\mathcal{U})$
	and $f$ is a monotonic function such that  $f[s (\mathcal{\hat{U}})] = J(\mathcal{\hat{U}})$ and $f = \argmin_{f} \gamma_f$, where $\gamma_f = \max_{\mathcal{U}} \vert J(\mathcal{U})-f[s(\mathcal{U})]\vert$. \label{def:tBSPerror}
	\end{definition}
	 
	In particular, $f$ for which \eqref{eq:error} is zero, corresponds to \tbsp being action consistent, i.e. $\mathcal{\hat{U}} = \mathcal{U}^{\star}$, and when also $\gamma_f = 0$, simplified representation preserves action order too.
	In the case $s = s_{ST}$, we can select $f(s) = -s + const.$ to quantify the error \eqref{eq:error} since the topological metric $s_{ST}$ is designed as negative of the entropy $J$ up to a constant when approximation error is small (see \cite{Kitanov18icra}). However, $f$ for  $s_{VN}$ is more complicated to determine and we leave this aspect to future investigation. Instead, we show empirically in realistic simulations that such function $f$ exists for which the error $\epsilon$ is small.
	
	Yet, as in \cite{Elimelech17isrr}, calculating $\epsilon(J, s)$ is essentially equivalent to solving the original problem. Therefore,  a key aspect will be to provide online performance guarantees by developing sufficiently tight bounds on $\epsilon(J, s)$ that can be evaluated online. One can then resort to topological BSP to drastically reduce computational cost while carefully monitoring a conservative estimate on the sacrifice in performance, that would be provided by the bound on $\epsilon(J, s)$. Another perspective of using this bound is to guarantee global optimality of anytime \tbsp algorithm we proposed in \cite{Kitanov18icra}. In that approach, actions were ranked by a topological metric and the objective function was evaluated sequentially from best to worst. Having bounds on \tbsp error would provide a stopping condition for action consistent \tbsp, similar to the action elimination scheme proposed in \cite{Elimelech17isrr} and its application to belief sparsification.


	
	

\subsection{Entropy bounds in BSP}
In BSP, we have to consider multiple path realizations from different controls, with greater variety in both topology of factor graphs and other non-topological factors that influence estimation accuracy, e.g non-fixed geometry and different path lengths. 
In \cite{Kitanov18icra} we  developed a topological metric $s_{ST}$ \eqref{eq:STsig} for \tbsp. Here, we show how to provide global optimality guarantees of \tbsp based on this metric.

For a control action $\mathcal{U}$ corresponding to robot's poses $X_{k+L} = \{x_1, ..., x_{k+L}\}$ each of dimension $\kappa$, the entropy of the future posterior belief $b[X_{k+L}]$ can be written using Eq.~\eqref{eq:ObjFunctionSep} and the fact $\text{ln}\lvert \Sigma(X_{k+L})\rvert = -\text{ln}\lvert I(X_{k+L})\rvert$
\begin{align}
J(\mathcal{U})={(k+L)\kappa}/{2}\ \text{ln} (2\pi \text{e}) - {1}/{2}\ \text{ln}\lvert I(X_{k+L})\rvert. \label{eq:entropy}
\end{align}
Notice that the number of graph nodes $\lvert V \rvert = n = k+L+1$.
Using Lemma 2 and Lemma 3 from \cite{Khosoussi15}, for posterior belief $b[\mathcal{P}^o] = \prob{X_{k+L}|\mathcal{U}_{0:k+L-1}}$ corresponding to an odometry factor graph whose topological graph $G^o$ is a tree, it follows
\begin{align}
\ln \text{det} I_o(X_{k+L}) & = \underbrace{\tau(G^o)}_{0} + \ln(\sigma_p^{-4}\sigma_\theta^{-2})^{k+L} \nonumber\\
& = (k+L) \ln \text{det}(\Omega_{\nu_{i,j}}). \label{detInfOdom}
\end{align}
Inequalities \eqref{eq:detInfBounds} and Eqs. \eqref{eq:entropy} and \eqref{detInfOdom}
give the entropy bounds
$\mathcal{L}\mathcal{B} [ J ( \mathcal{U} ) ] \leqslant J ( \mathcal{U} )
\leqslant \mathcal{U}\mathcal{B} [ J( \mathcal{U} )],\ \text{where}$
{\footnotesize
\begin{align}
& \mathcal{U}\mathcal{B} [ J ( \mathcal{U} ) ] = \dfrac{(n-1)\kappa}{2}  \ln
( 2 \pi e ) -\dfrac{1}{2} \left[ 3 \tau ( G ) + ( n-1 ) \ln | \Omega_{\nu_{i,j}} | \right] \nonumber \\
& =  -\dfrac{3}{2} \tau ( G ) - \dfrac{n-1}{2} [ \ln | \Omega_{\nu_{i,j}} | - \ln ( 2 \pi e)^{\kappa} ] =-s_{ST}(\mathcal{U}) \label{eq:UpperBound}
\end{align}}
Similarly, for the lower bound we get
{\footnotesize
\begin{align}
\mathcal{L}\mathcal{B} [ J ( \mathcal{U} ) ]  &=  -s_{ST} (\mathcal{U} ) -1/2 \ln | \tilde{L} + \Psi(\mathcal{U}) I_{n-1} | + 1/2\ \tau (G ) = \ldots \label{eq:LB_a}\\
& -s_{ST} ( \mathcal{U} ) -1/2 [\ln | \tilde{L} + \Psi(\mathcal{U})
I_{n-1} | - \ln | \tilde{L} | ]= \ldots \label{eq:LB_b} \\
&-s_{ST} ( \mathcal{U} ) -1/2  \ln| I_{n-1} + \tilde{L}^{-1} \Psi(\mathcal{U}) |. \label{eq:LB_c}
\end{align}
}
In eq. \eqref{eq:LB_b} we used the Matrix tree theorem that states $t(G) = | \tilde{L} |$, and in eq. \eqref{eq:LB_c} Shur's determinant lemma. From eq. \eqref{eq:LB_c} we see that when $\Psi \rightarrow 0$, the entropy goes to the BSP topological metric and solely depends on the belief's topology and path length. 
Otherwise, we have to account for the graph embedding in the metric space as well, as it appears in the scalar $\Psi$ of the second term in the above equations. 
Now, given all candidate control actions, the following theorem provides the bounds on the \tbsp error as defined by Definition \ref{def:tBSPerror}.

\begin{theorem}
For the error $\epsilon(J, s_{ST})$ of \tbsp, we can write
$$\epsilon(J, s_{ST})\leqslant \Delta J_{max},$$ where  $\Delta J_{max} = \mathcal{U}\mathcal{B} [ J ( \hat{\mathcal{U}} ) ] -
\underset{\mathcal{U}}{\min} \mathcal{L}\mathcal{B} [ J ( \mathcal{U} ) ].$
\end{theorem}
\begin{proof}
Rewrite $\Delta J_{max}$ as 
{\small
\begin{align*}
\Delta J_{max}  = \{ J ( \hat{\mathcal{U}} ) &+ \underbrace{\mathcal{U}\mathcal{B} [ J (
\hat{\mathcal{U}} ) ] -J ( \hat{\mathcal{U}} )}_{\delta ( \hat{\mathcal{U}} )
\geq 0} \} +\\
\{ -J ( \mathcal{U}^{\ast} ) &+ \underbrace{J (
\mathcal{U}^{\ast} ) - \underset{\mathcal{U}}{\min} \mathcal{L}\mathcal{B} [ J
( \mathcal{U} ) ]}_{\delta ( \mathcal{U}^{\ast} ) \geq 0} \}.
\end{align*}}
Therefore, $\delta = \delta ( \hat{\mathcal{U}} )+\delta ( \mathcal{U}^{\ast} ) \geq 0$ and $\Delta J_{max} = J (\hat{\mathcal{U}} ) -J ( \mathcal{U}^{\ast} ) + \delta \geq 0$ because $\mathcal{U}^{\ast}$ is the minimum of the entropy. Also, $\mathcal{U}\mathcal{B} [ J ( \hat{\mathcal{U}} )]
=-s_{ST} ( \hat{\mathcal{U}} )$ is the minimum upper bound of all actions since we select action in \tbsp according to eq. \eqref{eq:selectedCtrl}. We know that the optimal action must have entropy below this value, otherwise the selected action would be better which is contradiction by itself. Then,
$\Delta J_{\max} =J ( \hat{\mathcal{U}} ) -J ( \mathcal{U}^{\ast} ) + \delta =
| J ( \hat{\mathcal{U}} ) -J ( \mathcal{U}^{\ast} ) | + \delta \Rightarrow | J
( \hat{\mathcal{U}} ) -J ( \mathcal{U}^{\ast} ) | = \varepsilon ( J,s_{ST} ) =
\Delta J_{\max} - \delta \leqslant \Delta J_{\max}$.
\end{proof}

\subsection{Efficient calculation of entropy bounds}
The upper bound of the entropy is already determined by the topological metric as can be seen from eq. \eqref{eq:UpperBound}. However, calculating the lower bound requires an additional cost due to the second term in equations \eqref{eq:LB_a}-\eqref{eq:LB_c}. Calculating it requires evaluating the determinant of a sparse matrix $M = \tilde{L} + \Psi
I_{n-1} \in \mathbb{R}^{n-1\times n-1}$.

Another idea is to find some fast method for limiting the determinant of $M$ from above to replace the second term in eq. \eqref{eq:LB_a} that will not introduce a big difference in tightness of the lower bound.

\paragraph*{{Exact lower bound}}
A direct approach performs from scratch some sparse matrix factorization of $M$, e.g. Cholesky factorization $M = R R^T$ where $R$ is the lower triangular matrix, from which then it is easy to calculate its determinant. 
However, this approach, to be efficient, still requires finding a good fill-reducing permutation $P_M$ of $M$. The problem of finding the best ordering is an NP-complete problem \cite{Yannakakis1981} and is thus intractable, so heuristic methods are used instead.
However, we notice that some calculations from the topological metric can be re-used. In particular, since $M$ differs from $\tilde{L}$ only in diagonal elements, they both have the same sparsity pattern. Therefore, if $P_{\tilde{L}}$ is the best fill-reducing permutation of $\tilde{L}$ (already found for determining $\lvert \tilde{L} \rvert$), it can be re-used for calculation of the lower bound, i.e. $P_M=P_{\tilde{L}}$.

\paragraph*{{Hadamard bound}}
Since $\tilde{L} $ is a reduced graph Laplacian, it is symmetric positive definite (SPD), and  because also $\Psi > 0$, the matrix $M$ is SPD. For large values of $\Psi$, the matrix $M$ becomes strongly diagonally dominant and Hadamard inequality gives a good approximation of its determinant, i.e.
\begin{equation}
| \tilde{L} + \Psi I_{n-1} | \leqslant \prod_{i=2}^{n} [ d ( i ) + \Psi ]. \label{eq:HadamardBound}
\end{equation}
Calculation of the right side of inequality \eqref{eq:HadamardBound} requires only multiplication of node degrees with added value of $\Psi$. Applying \eqref{eq:HadamardBound} to eq. \eqref{eq:LB_a}, we can get somewhat more conservative but faster to compute lower bound  
{\small
\begin{equation}
\mathcal{L}\mathcal{B} [J (\mathcal{U})] = -s_{\tmop{ST}} (\mathcal{U})+1/2 (\tau (\mathcal{U}) - \underset{}{\prod_{i =
		2}^n [d_{\mathcal{U}} (i) + \Psi (\mathcal{U})]}) \label{eq:LB_HB}
\end{equation}}

\subsection{Summary}
Algorithm \ref{alg:tbsp-perf} summarizes our proposed method and highlights its possible uses in BSP regarding desired performance specifications.
Performance guarantees can be either in the form of selected solution's entropy upper bound, i.e. guarantees on the accuracy, or in bounding the error of \tbsp with respect to the optimal solution. 
%
%
%
\begin{algorithm}                      
	\DontPrintSemicolon
	\KwIn{set of factor graphs $FG$ corresponding to control actions $\mathcal{U}$ } 
	\KwOut{approximate solution to the BSP $\hat{\mathcal{U}}$ and
	its performance guarantees or actions subset $A$} 
	\KwData{graph signature $s\in \{s_{ST}, s_{VN}, \hat{s}_{VN}\}$, uncertainty margin $\gamma$ (optional)} 
	
	$S = \emptyset$\;
	represent each $FG$ with a topological graph $G(\mathcal{U})$\;
	\ForEach{$\mathcal{U}$}{
		evaluate the topological metric $s(\mathcal{U}) \doteq s[G(\mathcal{U})]$\;
		$S \gets S \cup \left\lbrace \mathcal{U}, s(\mathcal{U})\right\rbrace$
	}
%
	select $\hat{\mathcal{U}} = \underset{\mathcal{U}}{\arg \max\ } s(\mathcal{U})$\;
	$\mathcal{U}\mathcal{B} [J (\hat{\mathcal{U}})] = - s_{\tmop{ST}}(\hat{\mathcal{U}})$\; \label{alg:selectUpperBound}
	
	\uIf {uncertainty margin $\gamma$ is given} {
	
		\uIf {$\mathcal{U}\mathcal{B} [J (\hat{\mathcal{U}})] < \gamma$} {
			\Return $\hat{\mathcal{U}}$
		}
		\Else {
			$A = sort(S)$ with respect to topological metric	\Return $A$
		}
	}
	\Else {			
		$\mathcal{L}\mathcal{B}_{\min} = \underset{\mathcal{U}}{\min} \mathcal{L}\mathcal{B} [J (\mathcal{U})]$ using eq. \eqref{eq:LB_HB}\;
		$\Delta J_{\max}$ = $\mathcal{U}\mathcal{B} [J (\hat{\mathcal{U}})] -
		\mathcal{L}\mathcal{B}_{\min}$\;
		$A = \left\lbrace \mathcal{U} \in S : \mathcal{L}\mathcal{B} [J (\mathcal{U})] < \mathcal{U}\mathcal{B} [J (\hat{\mathcal{U}})] \right\rbrace $\;
		\Return $\hat{\mathcal{U}}$, t-bsp error bound $\Delta J_{\max}$, $A$
	}
\caption{\tbsp with performance guarantees}          
\label{alg:tbsp-perf}                           
\end{algorithm}

The first form can be used when the maximum admissible path uncertainty is known at the planning time, e.g. for obstacle avoidance. In that case, one can get an answer  if a \tbsp solution satisfies the specification by ranking actions using very efficient $O(m |A|)$ topological metric $\hat{s}_{VN}$ and calculating only its entropy's upper bound (Alg. \ref{alg:tbsp-perf}, line \ref{alg:selectUpperBound}). If global optimality guarantees are required, the topological metric $s_{ST}$ needs to be calculated for all actions, so it is currently the only reasonable choice as a graph signature. 
In the first usage, if uncertainty specification is not met by \tbsp one can still use its ranked actions set $A$ in anytime algorithm as we proposed in \cite{Kitanov18icra}. Also, if an optimal solution needs to be found, we can use \tbsp to eliminate suboptimal actions.

\section{Results}
\label{sec:results}

\begin{figure}
	\centering
	\subfloat[S1 candidate paths on top of PRM]{
		\includegraphics[width= 0.48\columnwidth]{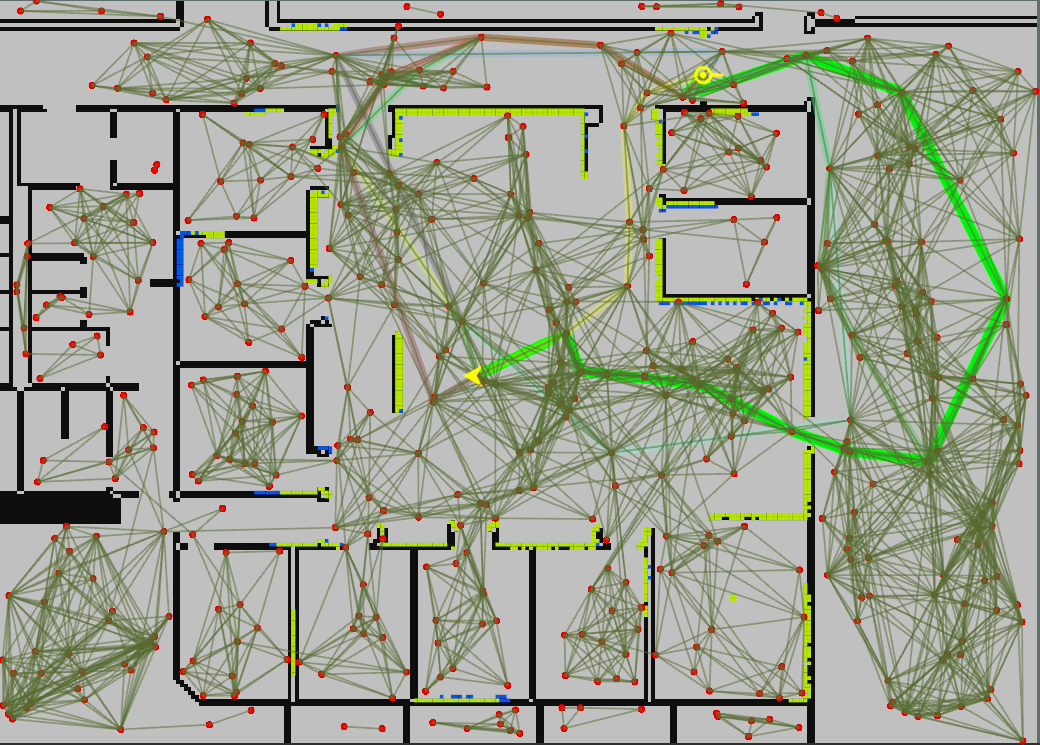}\label{fig:PRM}}
	\subfloat[S2 candidate paths]{
		\includegraphics[width= 0.48\columnwidth]{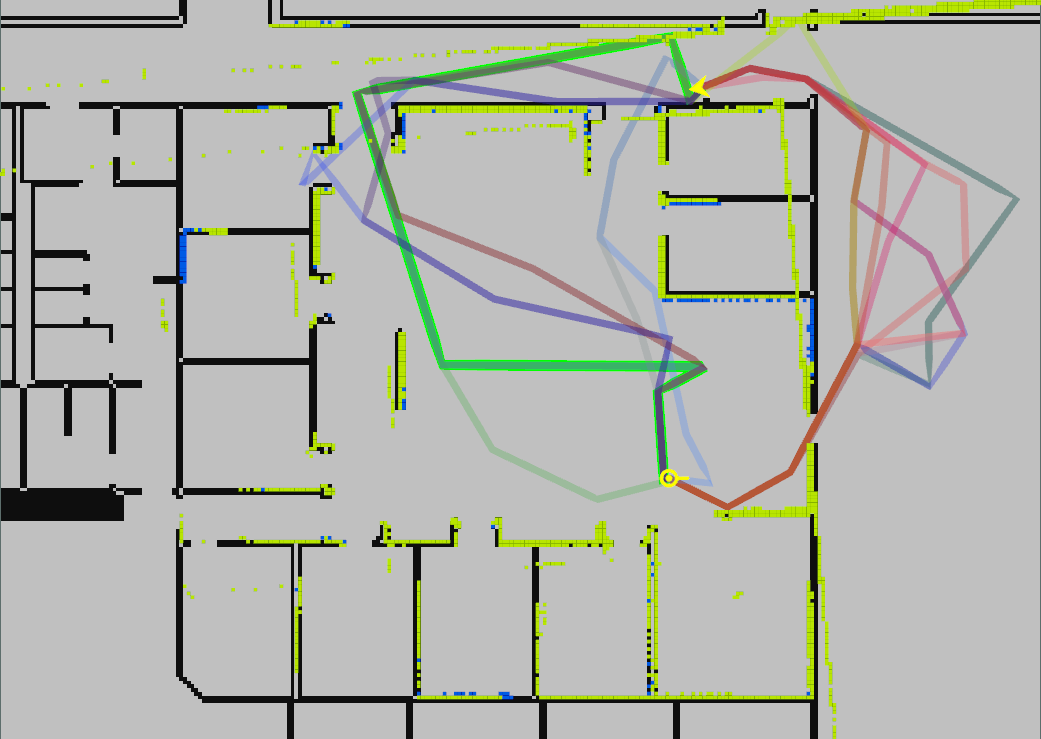}\label{fig:S2cand}}
	\caption[Gazebo scenario]{Gazebo scenario with robot's start pose marked with \includegraphics[height=6pt]{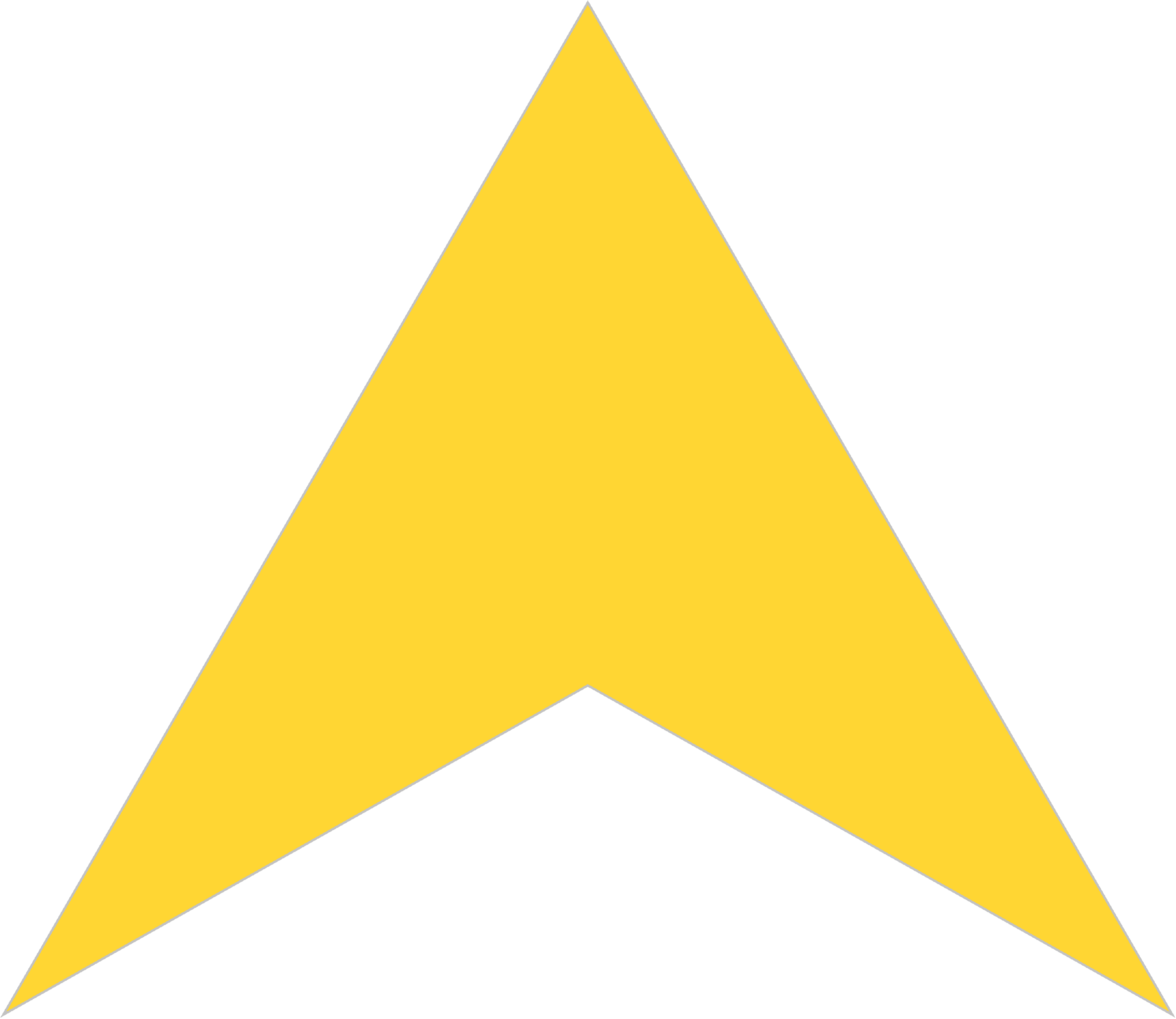} and goal pose with \includegraphics[height=6pt]{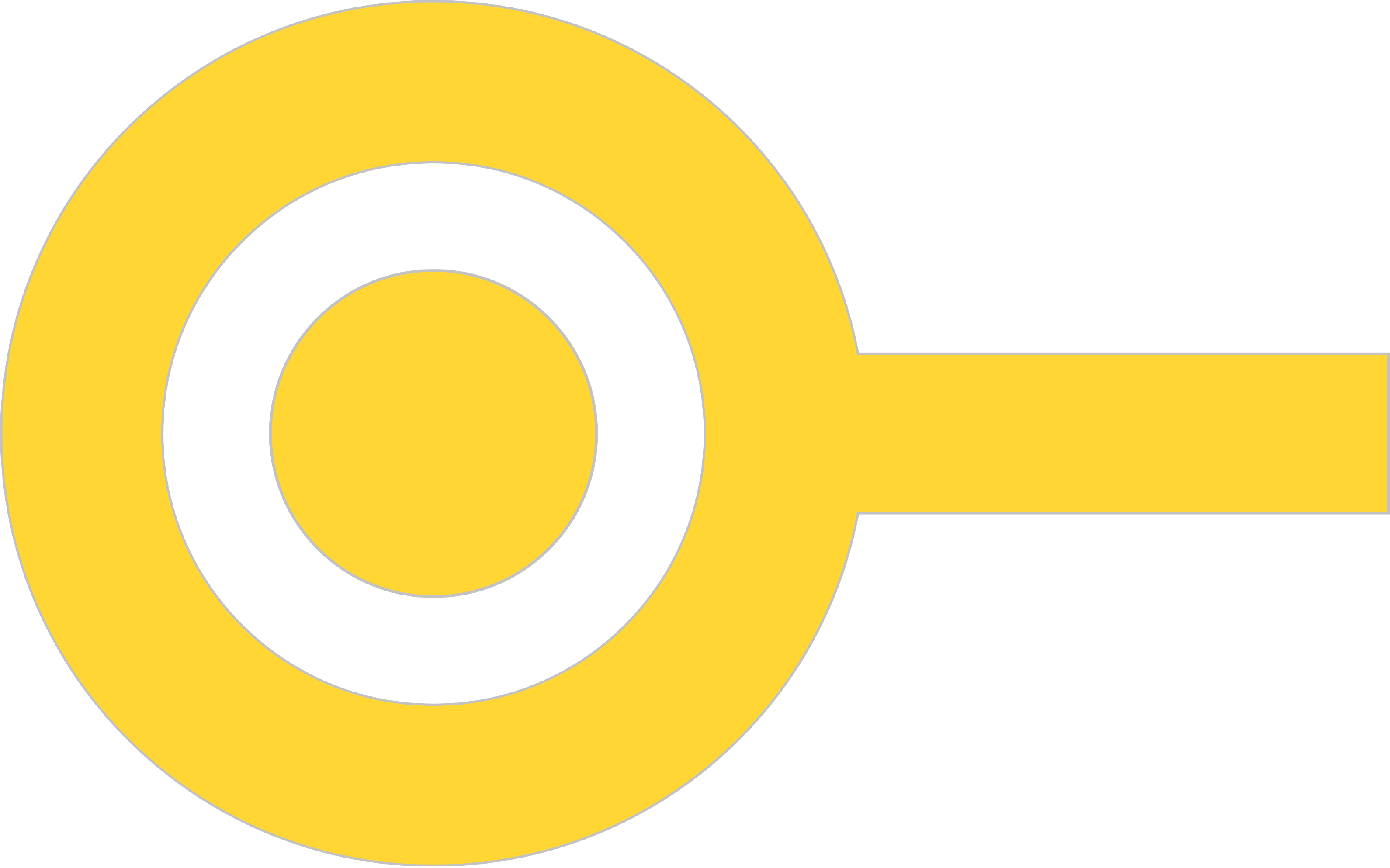} in each planning session.} 
	\label{fig:candidatePaths}
\end{figure}

We evaluated our approach in a Gazebo simulation of a single-robot active pose SLAM and in synthetic pose graphs optimization.
We studied empirically different topological metrics and their correspondences to information theoretic cost \eqref{eq:ObjFunctionSep}. The robot was performing two planning sessions with both exploration and exploitation trajectories considered to show influence of different candidate path lengths and  geometry in \tbsp. A probabilistic roadmap (PRM) \cite{Kavraki96tra} was used to discretize the environment and generate the roadmap (Fig. \ref{fig:PRM}) while \textit{k}-diverse shortest path algorithm \cite{Voss2015} to generate topologically diverse candidate paths over it. Figure \ref{fig:candidatePaths} shows the considered scenario in Gazebo and the generated candidate paths for the robot in two planning sessions S1 (Fig. \ref{fig:PRM}) and S2 (Fig. \ref{fig:S2cand}). To demonstrate the effect of noise level on \tbsp, we simulated three cases. All relative pose measurements had standard deviation of orientation error $\sigma_{\theta}$ either 0.01, 0.035 or 0.085 rad while we kept the same standard deviation of the position error $\sigma_{p}$ = 0.1 m. This corresponds to three different values of $\xi \in \{0.1, 0.35, 0.85\}$ for each planning session.  
 
\begin{figure}
	\centering
	\subfloat[S1 worst action]{
		\includegraphics[width= 0.2\textwidth]{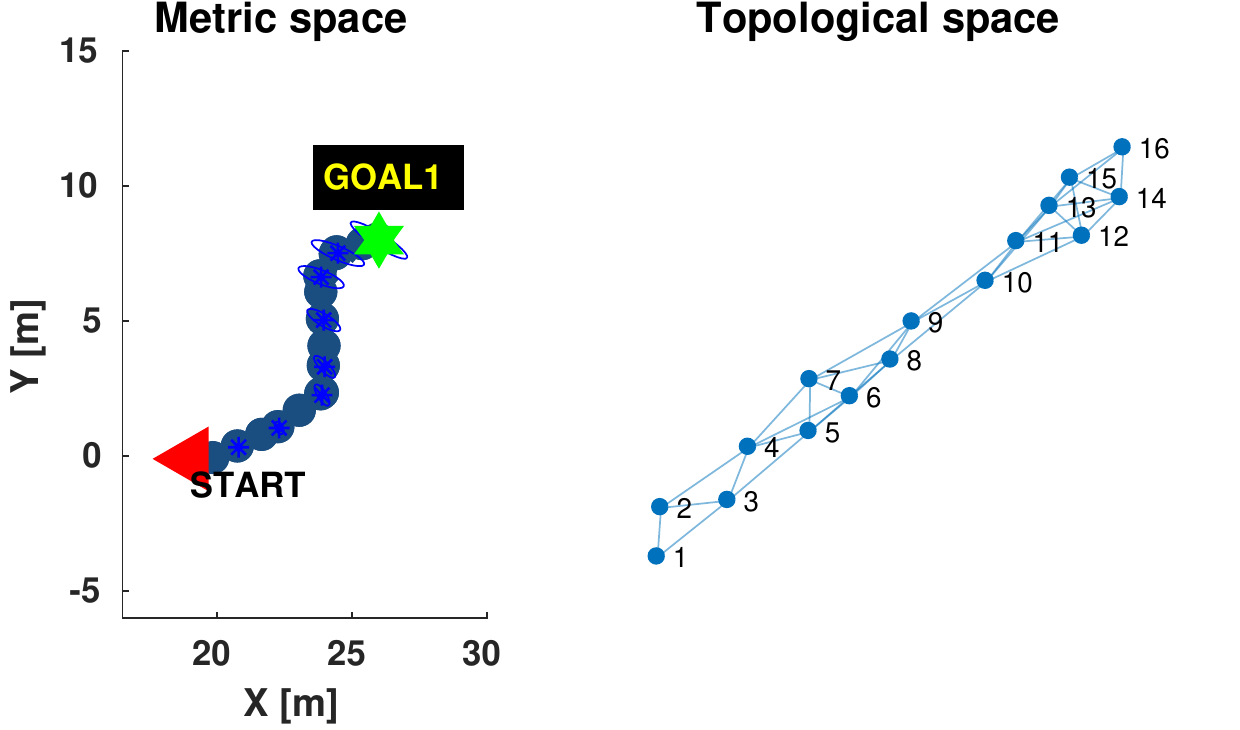}\label{fig:tExplorationL}}
	\subfloat[S1 best action]{
		\includegraphics[width= 0.23\textwidth]{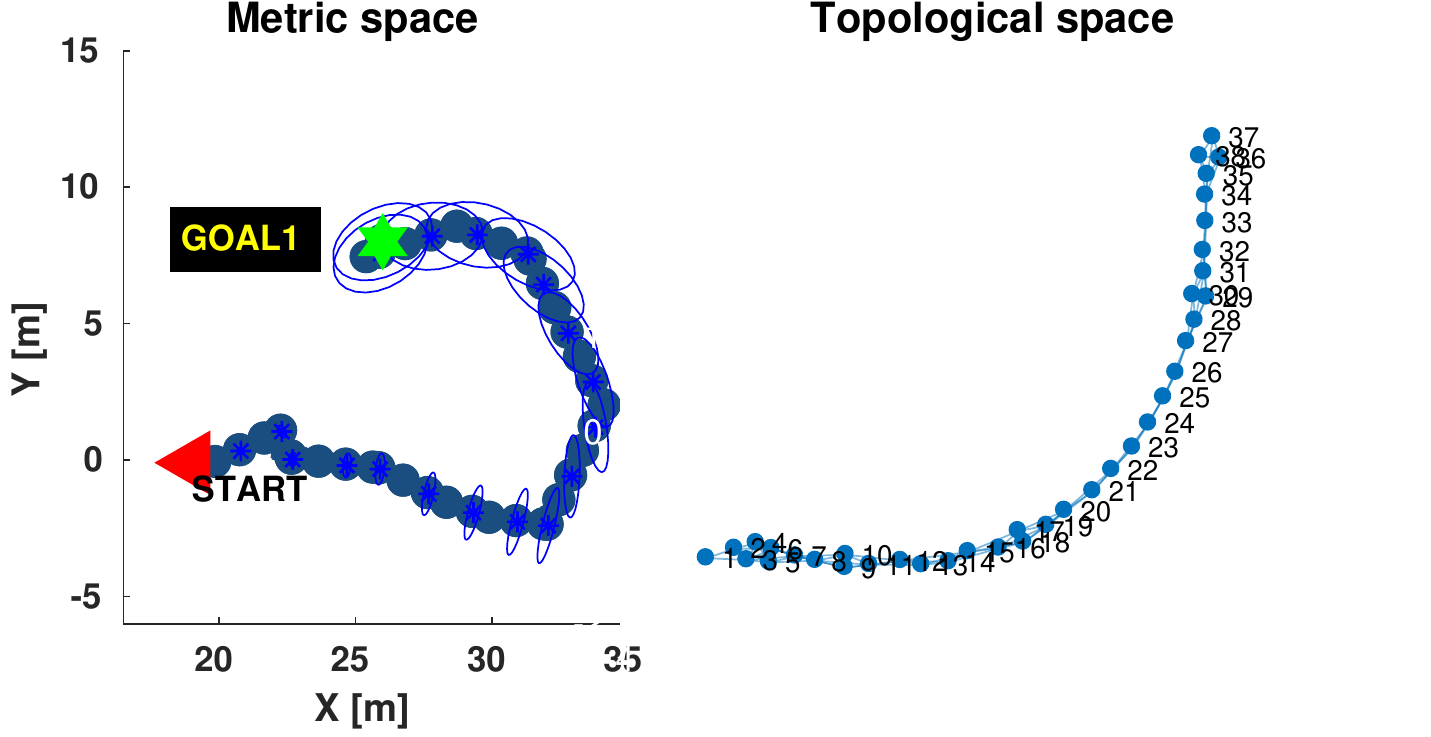}\label{fig:tExplorationH}}\\
	\subfloat[S2 worst action]{
		\includegraphics[width= 0.23\textwidth]{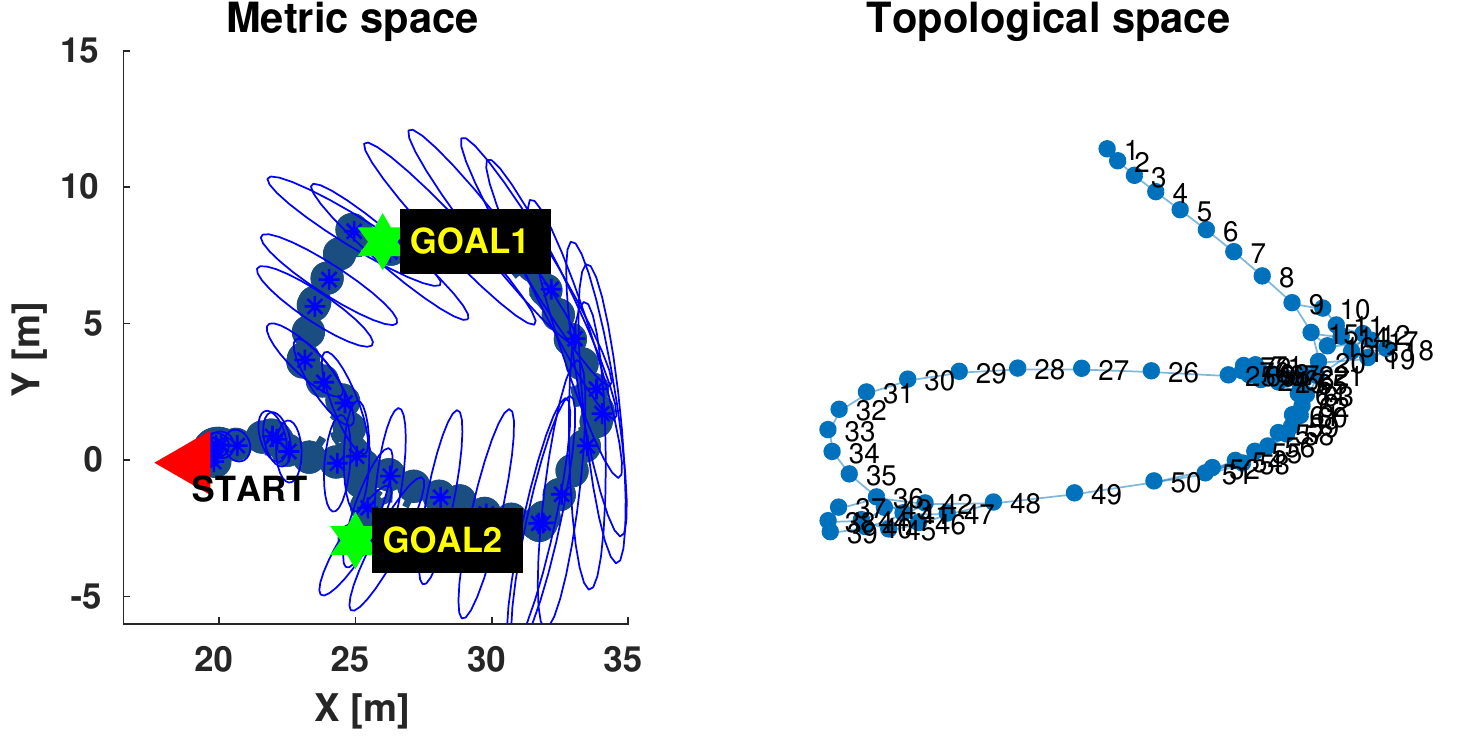}\label{fig:tExploatationL}}
	\subfloat[S2 best action]{
		\includegraphics[width= 0.23\textwidth]{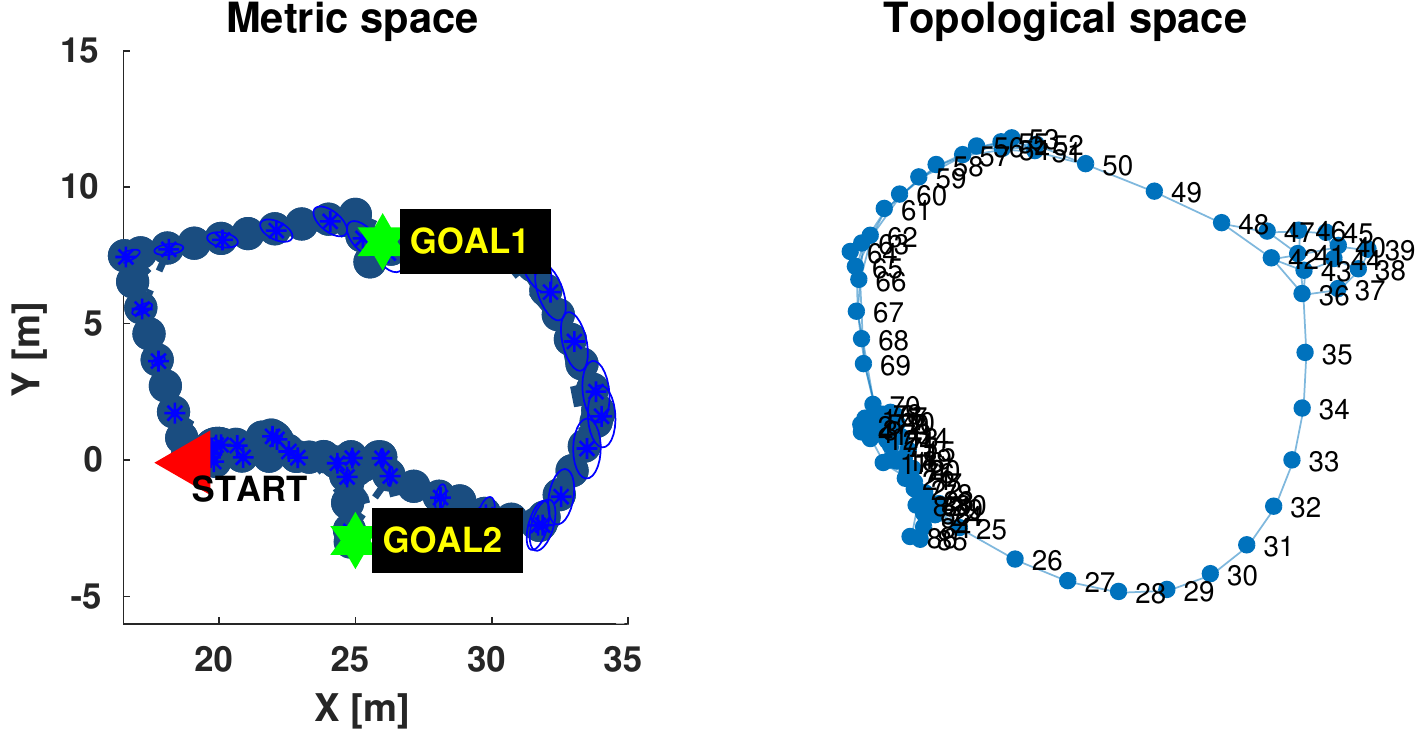}\label{fig:tExploatationH}}
	\caption{Trajectory uncertainty after optimization and topologies of the worst/best actions calculated by topological BSP.}
	\label{fig:topologies}
\end{figure}

\begin{figure*}
	\centering
	\subfloat[S1, $\xi = 0.1$]{\includegraphics[width= 0.16\textwidth]{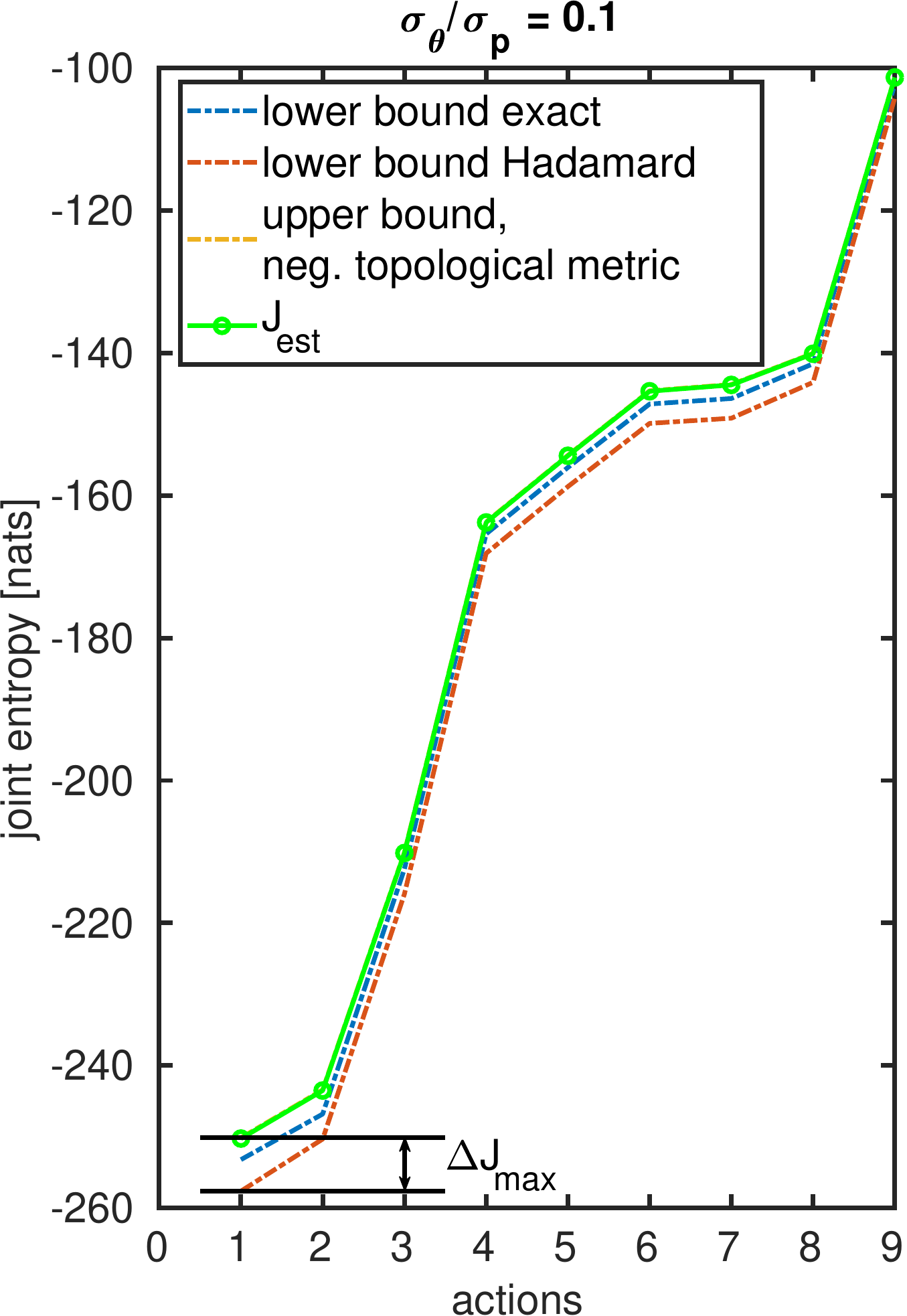}\label{fig:EntropyBoundsSession1_01}}
	\subfloat[S1, $\xi = 0.35$]{\includegraphics[width= 0.16\textwidth]{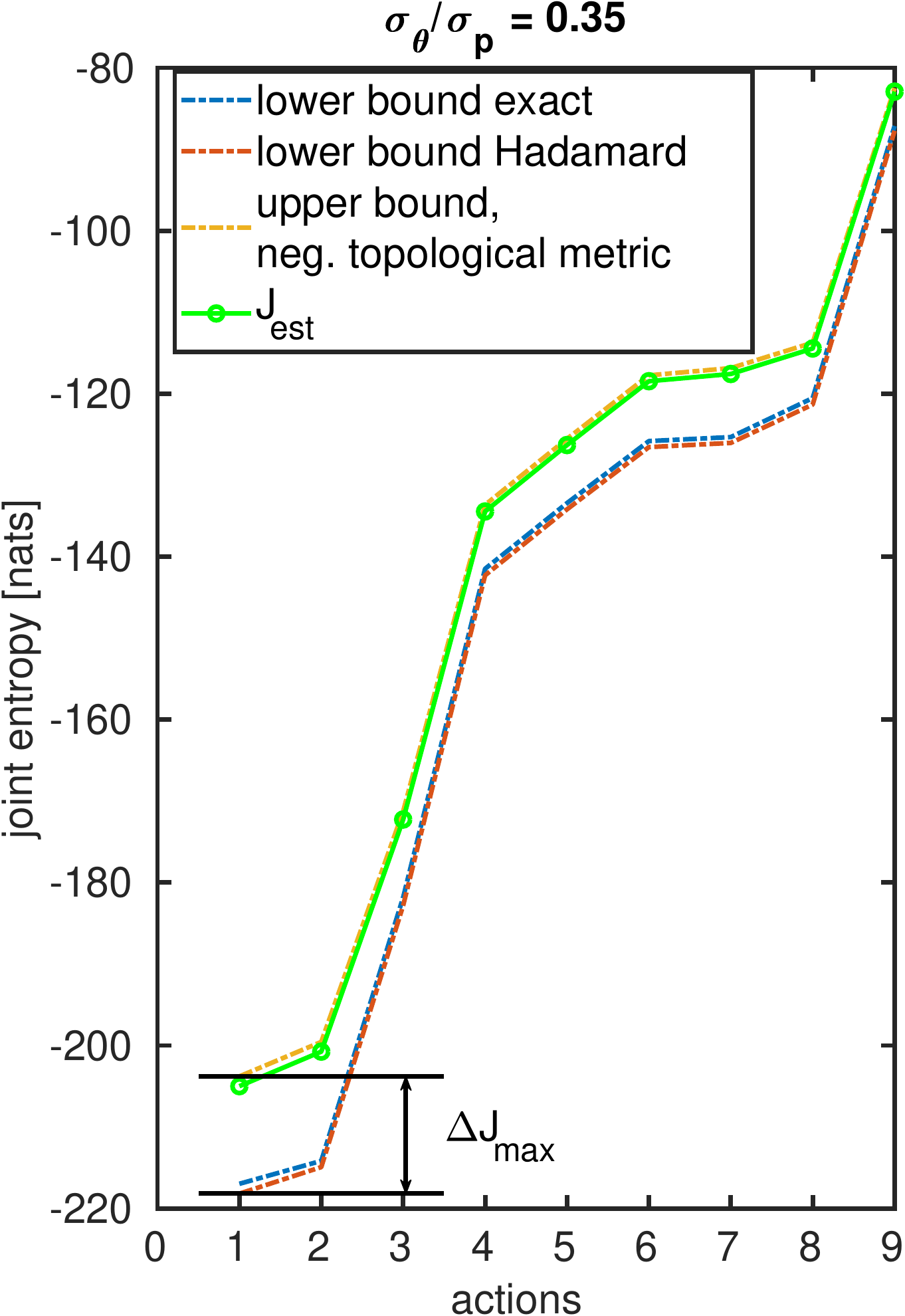}\label{fig:EntropyBoundsSession1_03}}
	\subfloat[S1, $\xi = 0.85$]{\includegraphics[width= 0.16\textwidth]{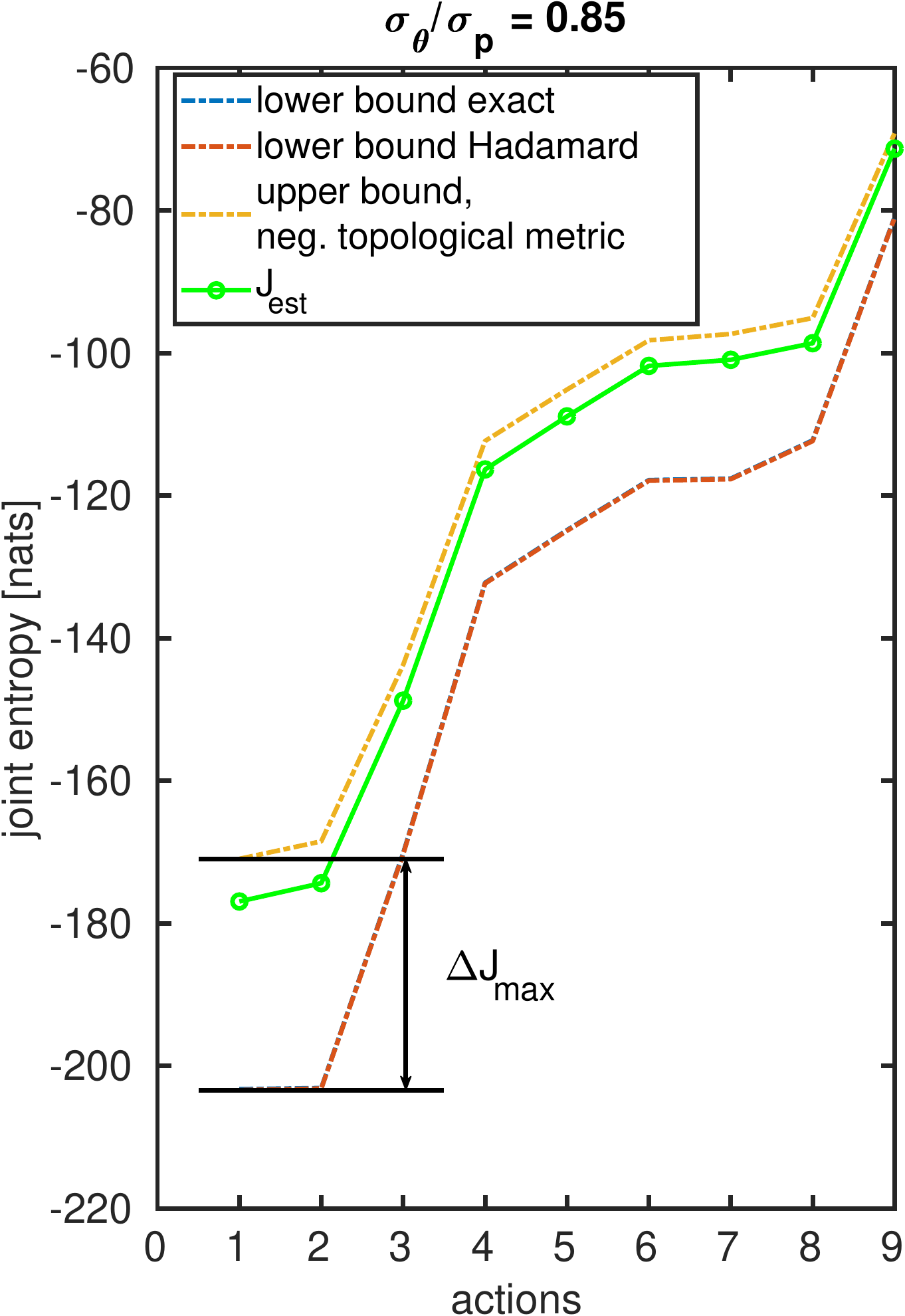}\label{fig:EntropyBoundsSession1_05}}
	\subfloat[S2, $\xi = 0.1$]{\includegraphics[width= 0.16\textwidth]{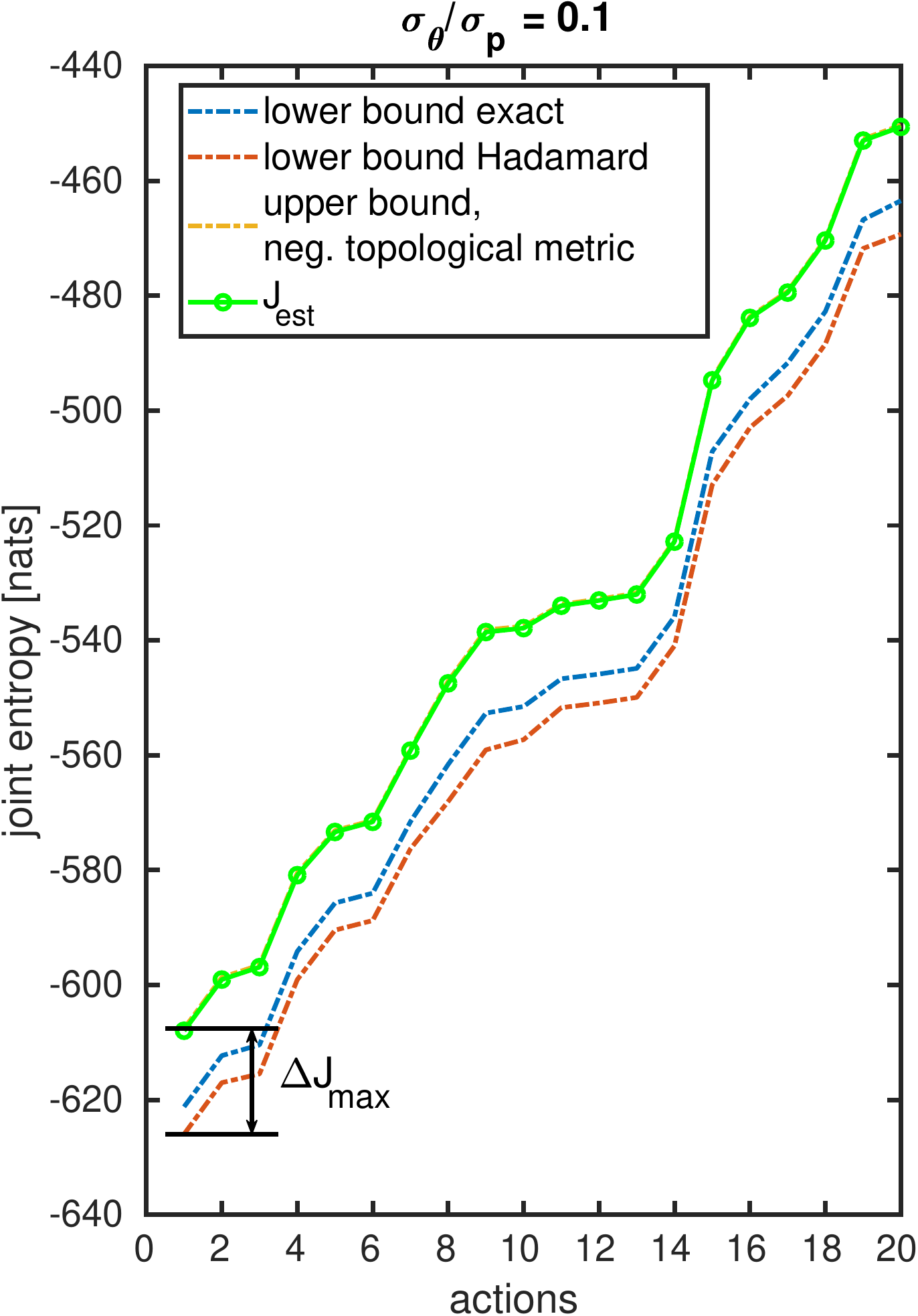}\label{fig:EntropyBoundsSession2_01}}
	\subfloat[S2, $\xi = 0.35$]{\includegraphics[width= 0.16\textwidth]{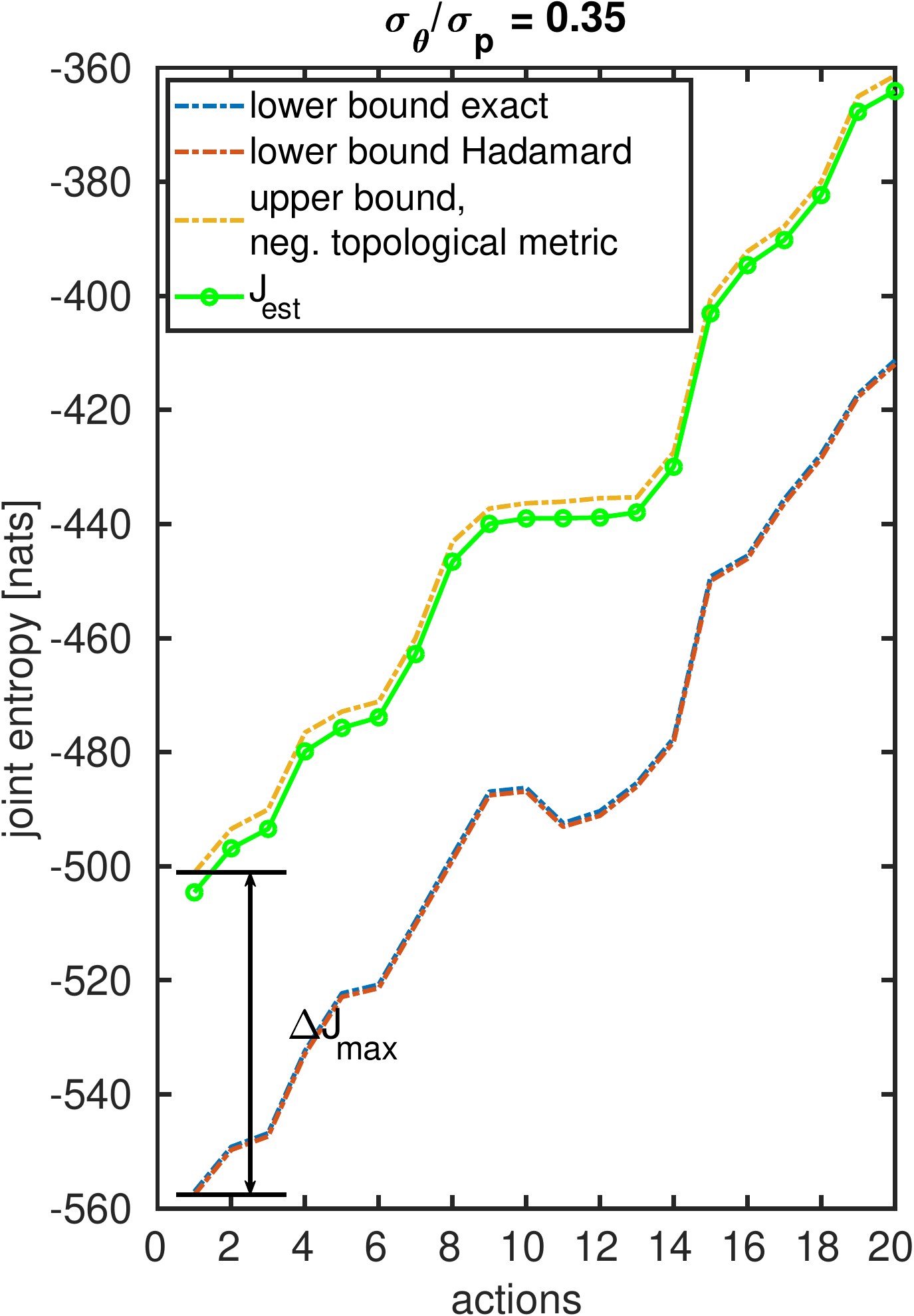}\label{fig:EntropyBoundsSession2_03}}
	\subfloat[S2, $\xi = 0.85$]{\includegraphics[width= 0.16\textwidth]{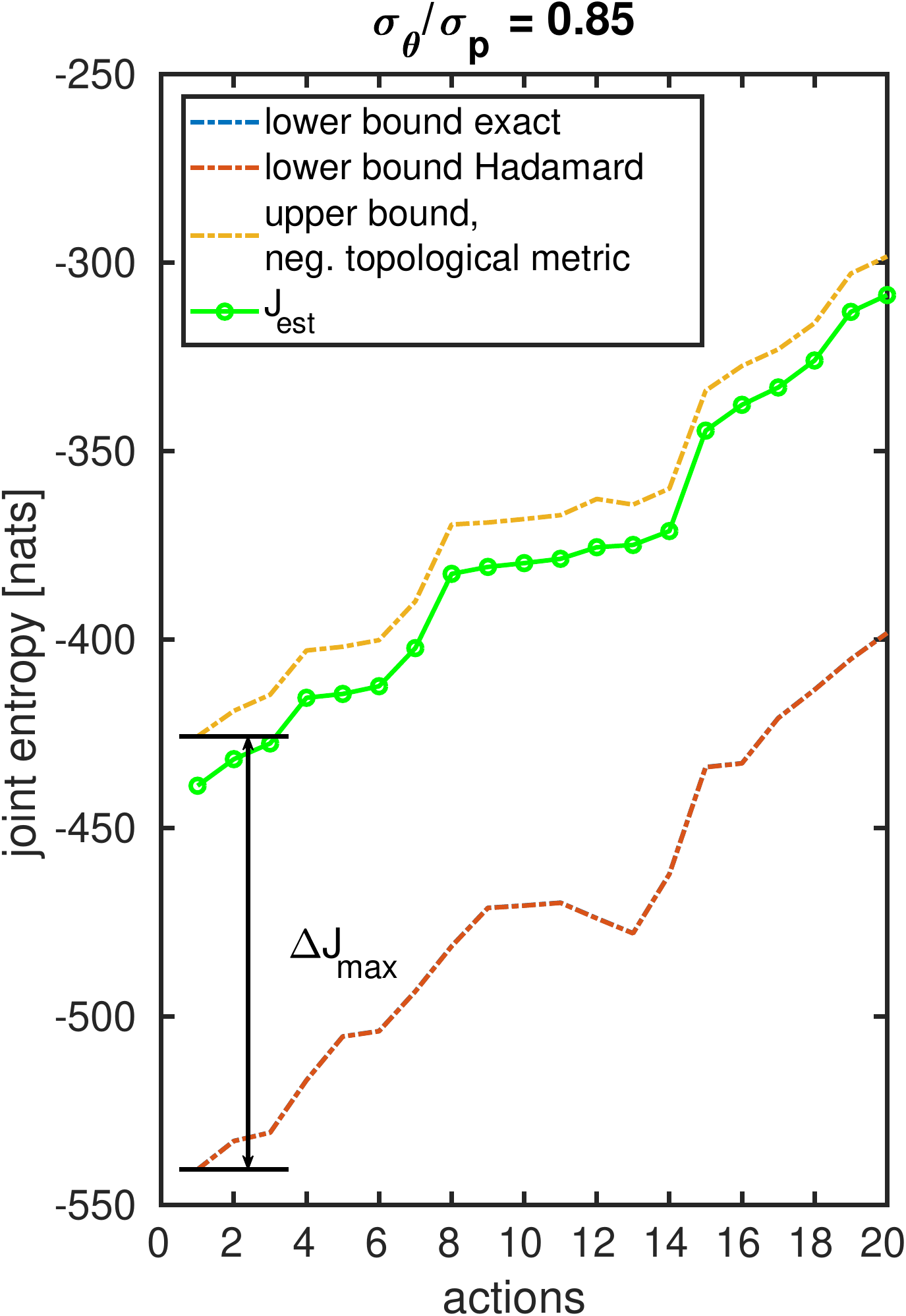}\label{fig:EntropyBoundsSession2_05}}
	
	\subfloat[S1, $\xi = 0.1$]{\includegraphics[width= 0.16\textwidth]{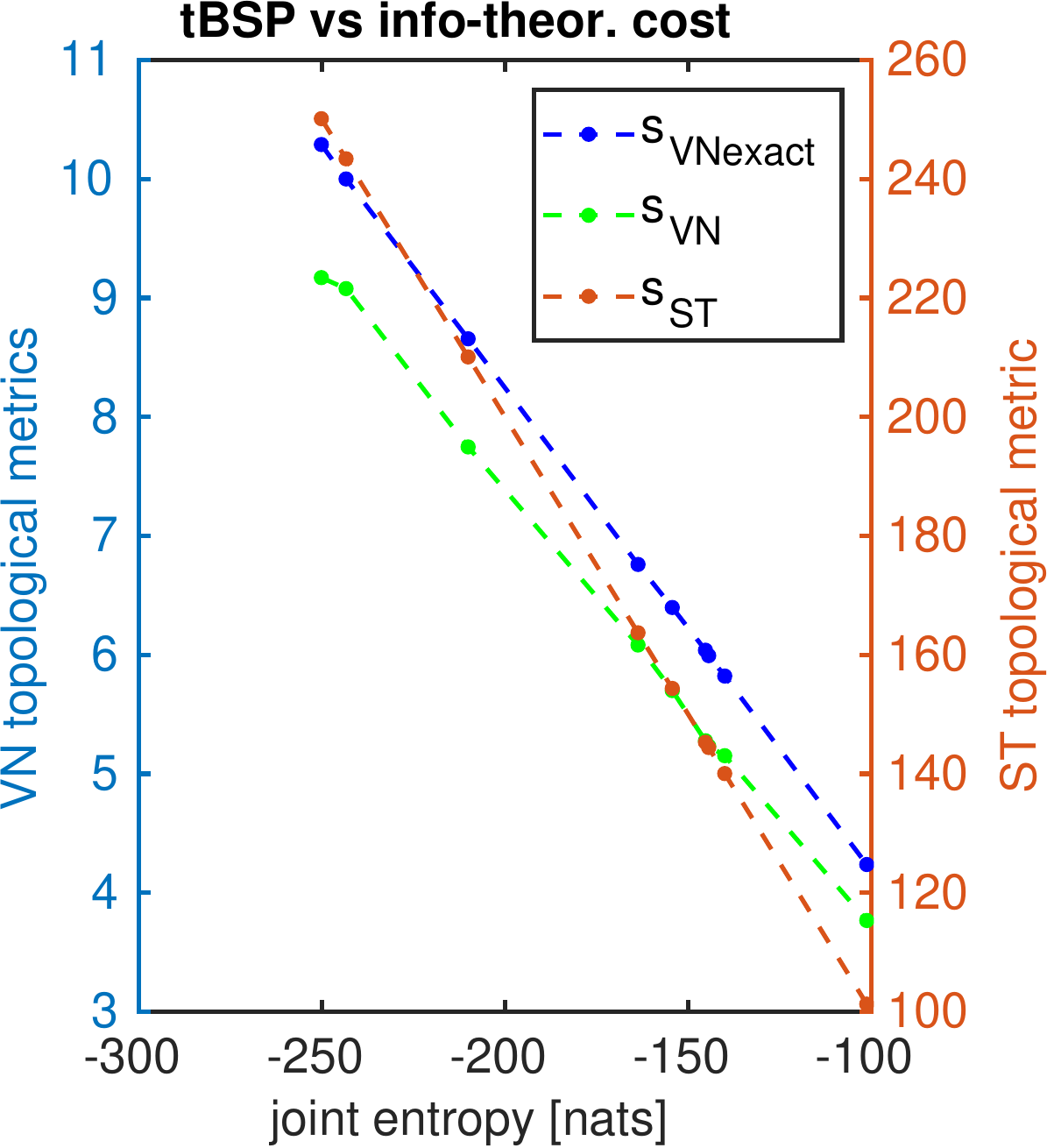}\label{fig:CorrSession1_01}}
	\subfloat[S1, $\xi = 0.35$]{\includegraphics[width= 0.16\textwidth]{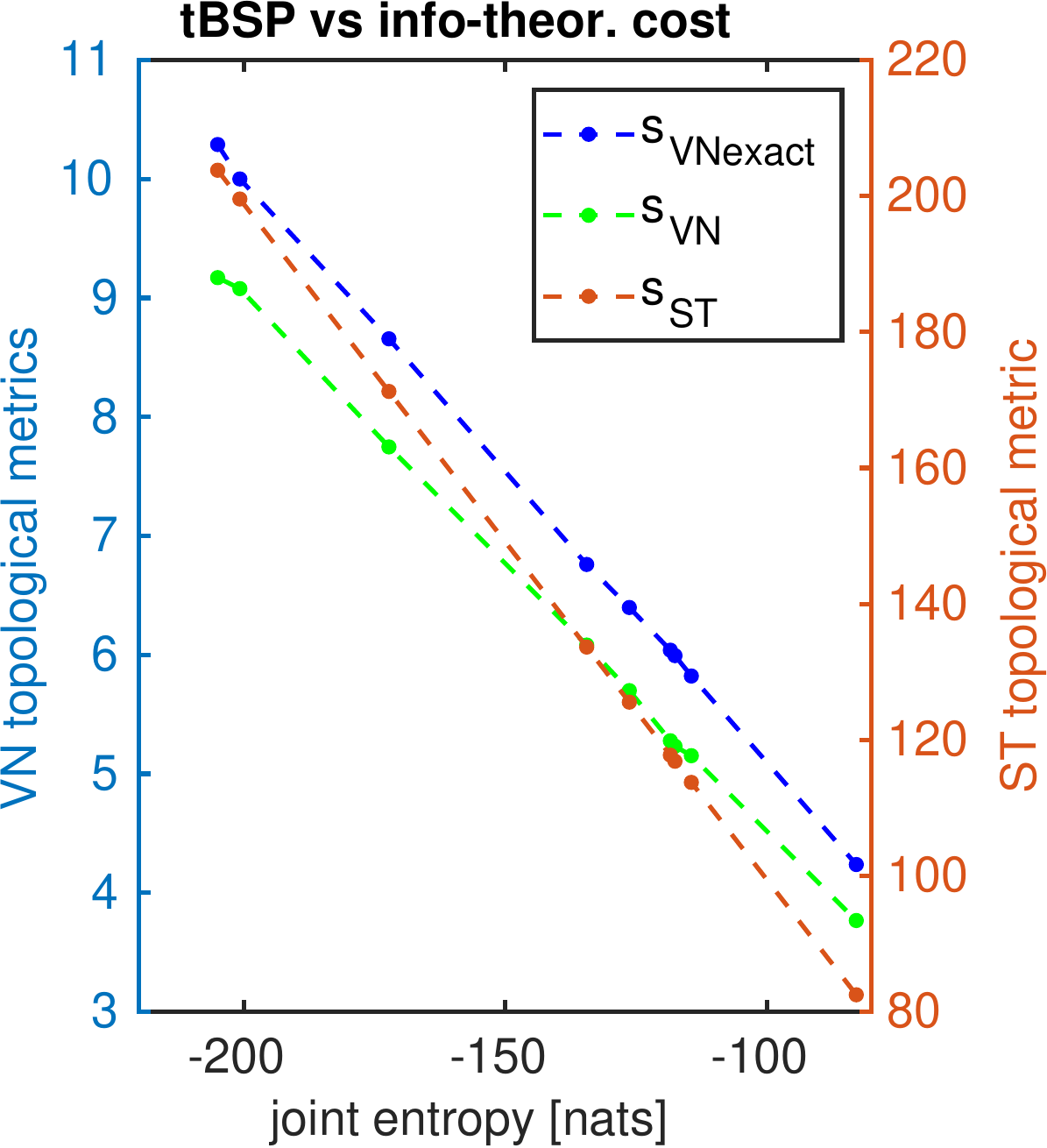}\label{fig:CorrSession1_03}}
	\subfloat[S1, $\xi = 0.85$]{\includegraphics[width= 0.16\textwidth]{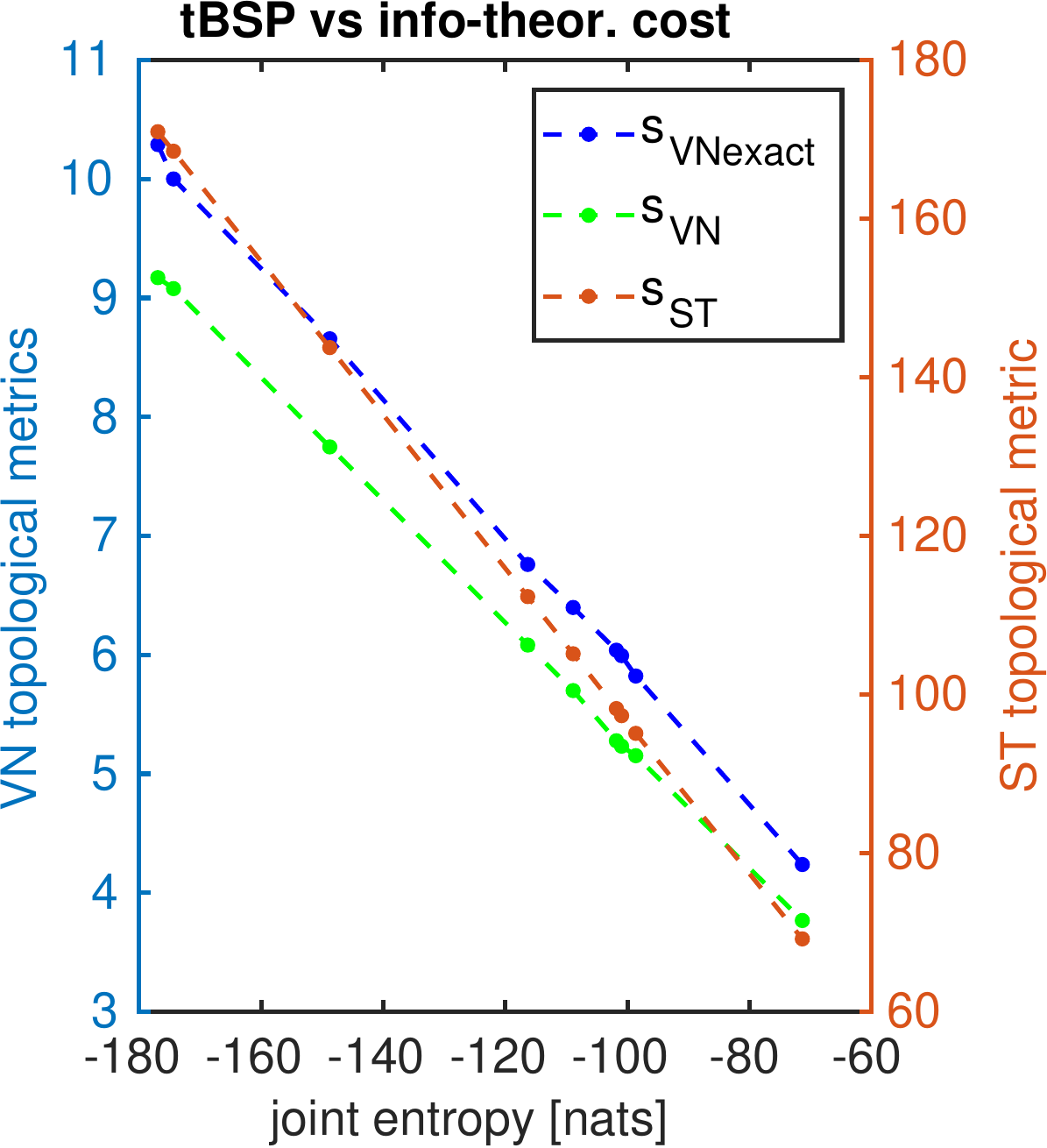}\label{fig:CorrSession1_05}}
	\subfloat[S2, $\xi = 0.1$]{\includegraphics[width= 0.16\textwidth]{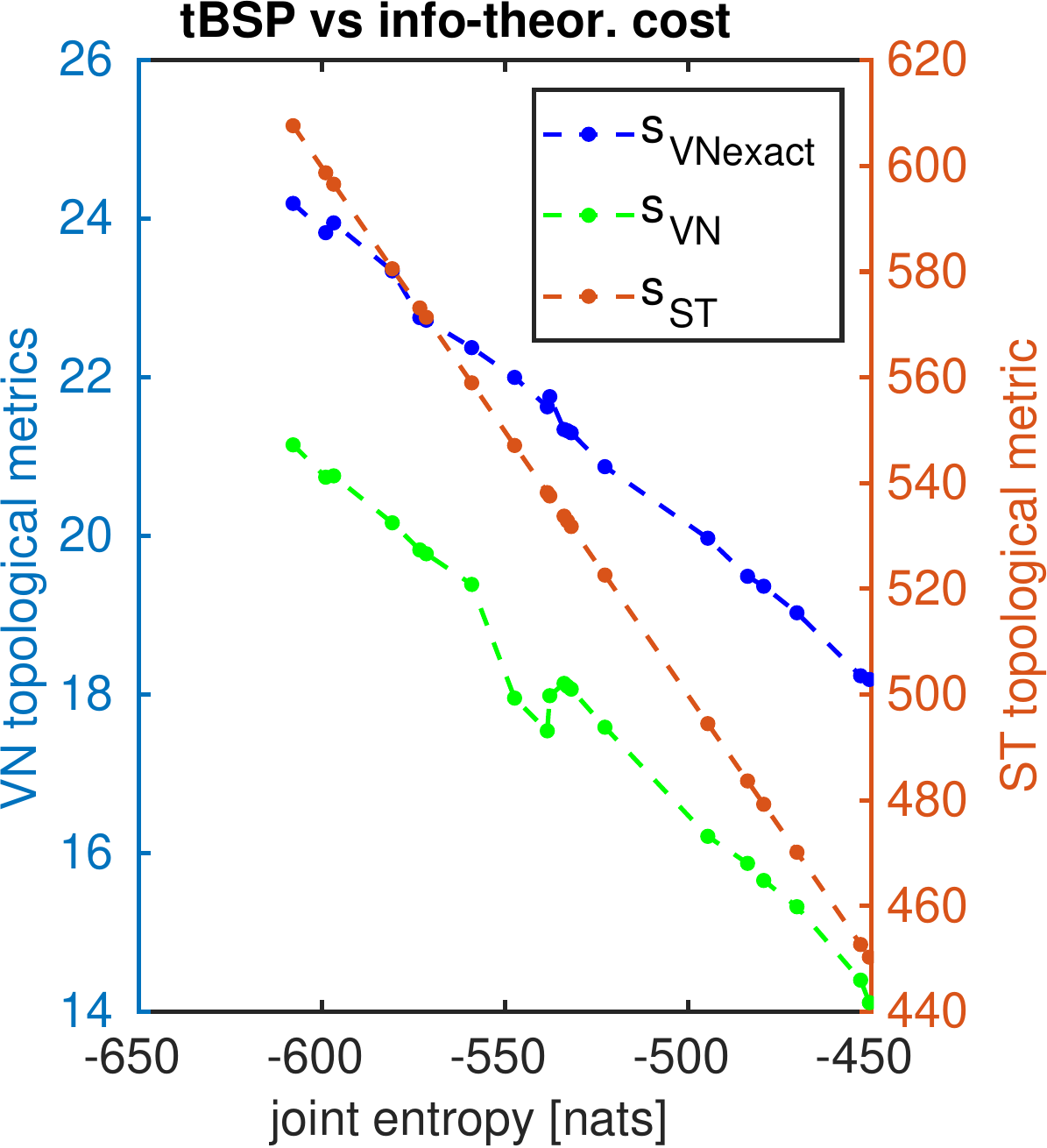}\label{fig:CorrSession2_01}}
	\subfloat[S2, $\xi = 0.35$]{\includegraphics[width= 0.16\textwidth]{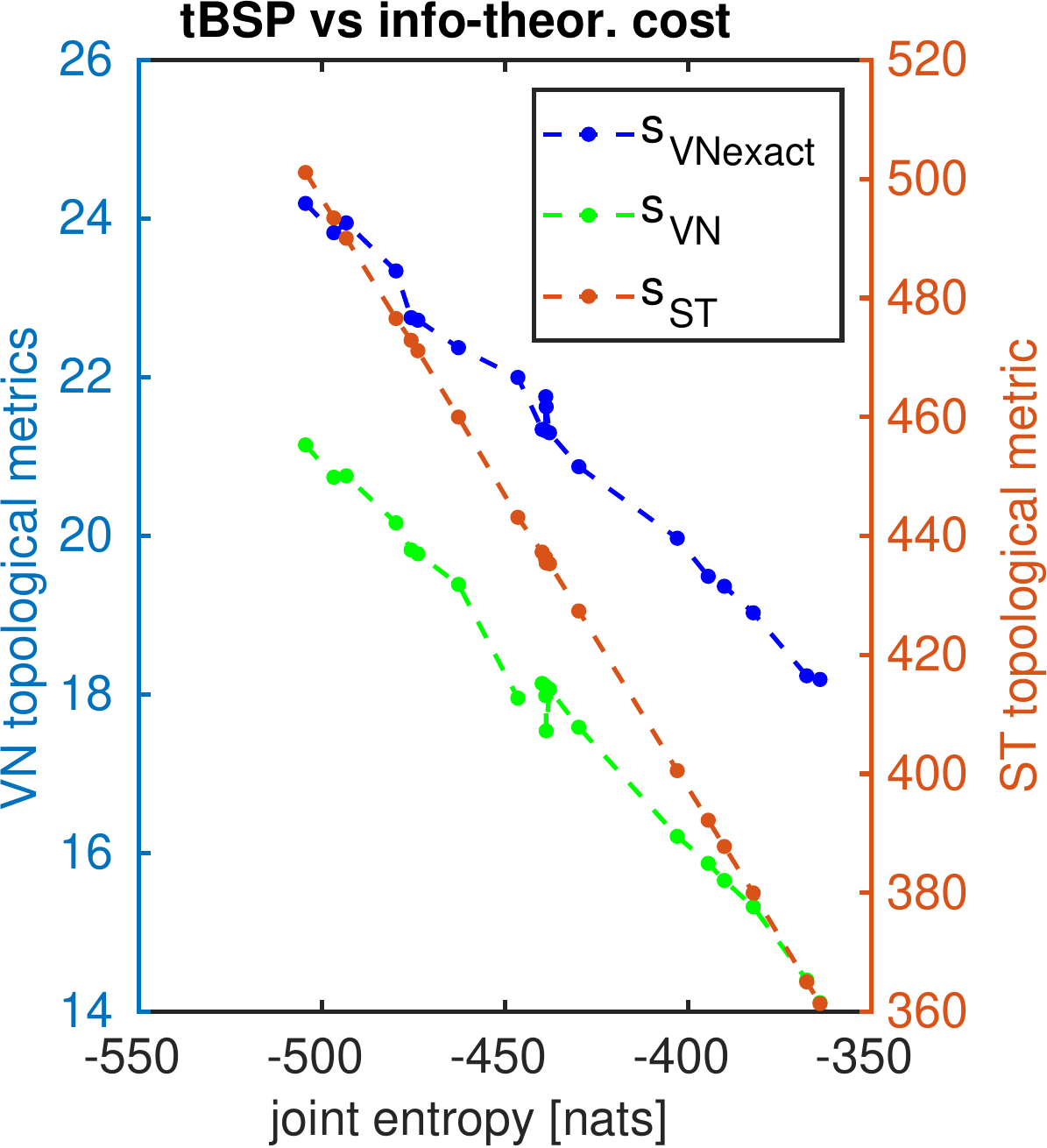}\label{fig:CorrSession2_03}}
	\subfloat[S2, $\xi = 0.85$]{\includegraphics[width= 0.16\textwidth]{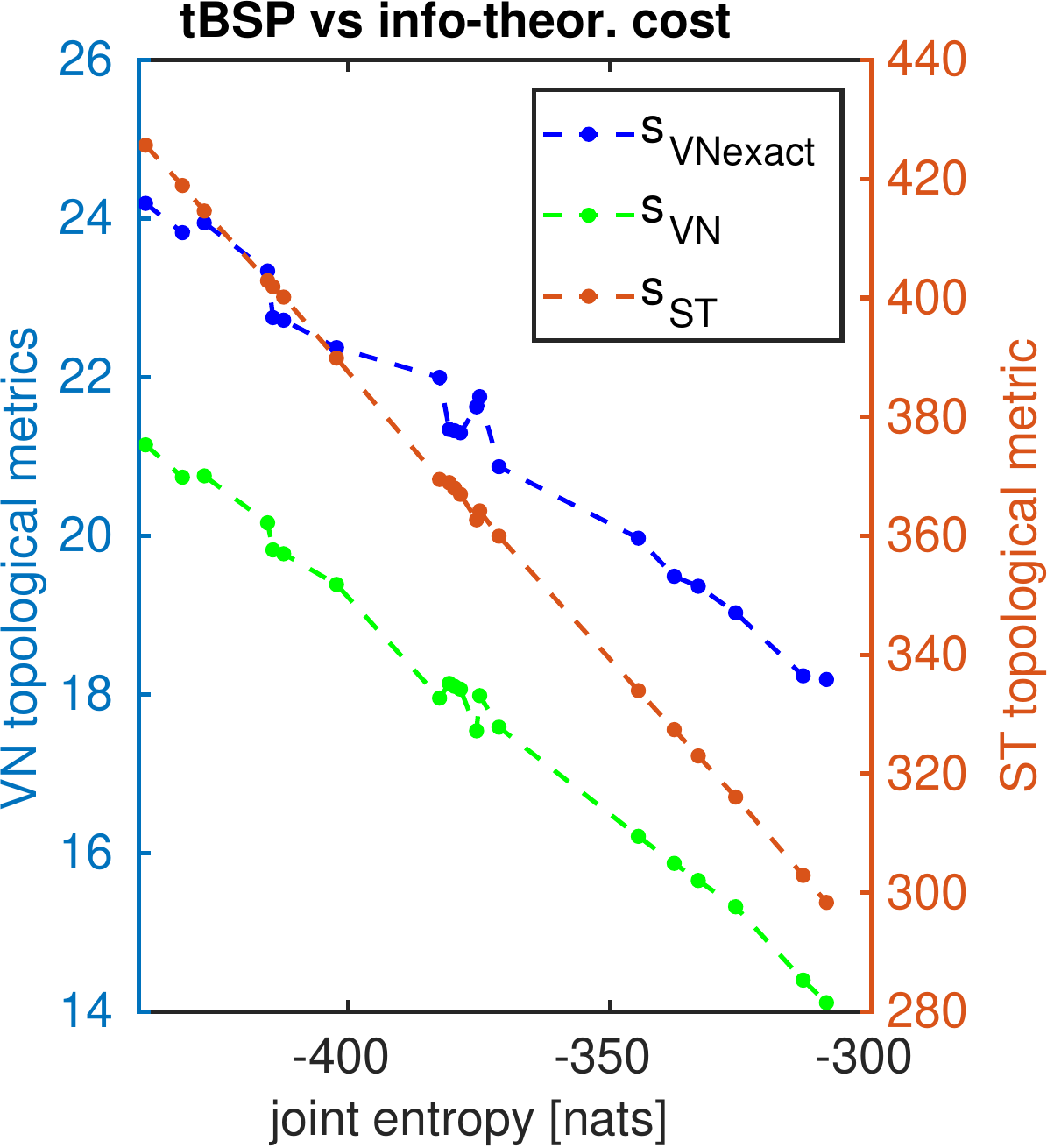}\label{fig:CorrSession2_05}}				
	\caption{Estimated entropy of posterior belief and its bounds determined by \tbsp (\textit{up}) and correlation of \tbsp and standard BSP (\textit{bottom}) for three different values of $\xi$ in planning sessions S1 and S2. Notice different $y$-axis on the correlation plots since only action trend consistency is important in BSP.}
	\label{fig:CorrAndEntropySession12}
\end{figure*}

The standard BSP inference problem was solved using GTSAM library \cite{Dellaert12tr}. During planning session S1,  the robot is performing mainly exploration as the environment was completely unknown in the beginning and its first goal is set far in the unknown area so its future posterior beliefs have topologies that resemble tree graphs, with loop closing edges between poses nearby in time. In planning session S2, the robot was instructed to return to the previously mapped area causing larger diversity among candidate actions. Topologies of the most and least complex graph of each planning session are shown in Fig. \ref{fig:topologies} together with their corresponding posterior belief. In both planning sessions \tbsp based on all proposed metrics was action consistent, i.e. correctly identified the best action which can be seen from Figs. \ref{fig:CorrSession1_01}-\ref{fig:CorrSession2_05}. A maximum of a topological metric is indeed a minimum of the joint entropy. Notice that it does not imply necessarily that the pose uncertainty will be the lowest at the goal or that the shortest trajectory will be selected as is evident form S1's solution (Figs. \ref{fig:tExplorationL}-\ref{fig:tExplorationH}) since currently the entropy/uncertainty of the entire system is considered. This opens one other relevant research direction where an uncertainty of a subset of state variables might be considered and its relation to topological information.
Among exploitation trajectories, \tbsp prefers the one with larger loop closings leading to highest information gain (see Figs. \ref{fig:tExploatationL}-\ref{fig:tExploatationH}).

\begin{figure}
	\centering
	\includegraphics[width=0.72\columnwidth]{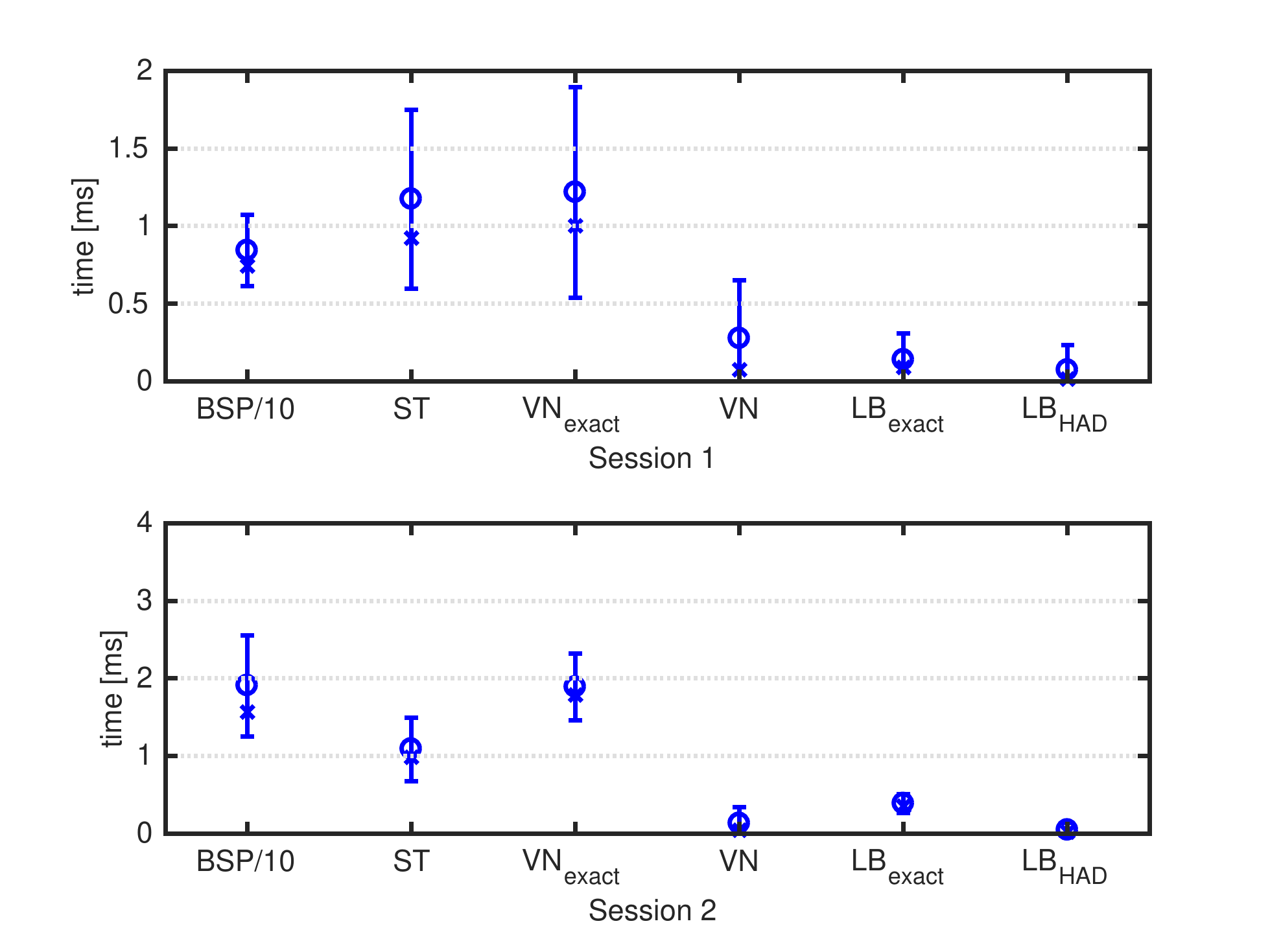}
	\caption{Computation time per candidate action in each planning session (circle represents the mean, 'x' the median and line interval one $\sigma$ confidence region) for standard BSP, \tbsp by metrics based on the number of spanning trees (ST), Von Neumann entropy (VN exact and approx.) and for bounds determination (exact and Hadamard). Notice that computation time of the standard BSP approach has been scaled by 10.}
	\label{fig:timePerAction}
\end{figure}

Our results in a realistic simulation experiment show the promise of the proposed approach for efficiently solving BSP.
All topological metrics of the posterior factor graphs are strongly correlated with the information-theoretic cost (Figs. \ref{fig:CorrSession1_01}-\ref{fig:CorrSession2_05}), yet
\tbsp based on an exact Von Neumann entropy ($s_{VN}$) and the number of spanning trees ($s_{ST}$) of a graph outperforms standard BSP by an order of magnitude in terms of time-complexity as shown in Fig. \ref{fig:timePerAction}. This is due to an operation in a topological (less dimensional) space instead of a metric state space. Their relative speed is similar as both of them require determining a graph spectrum of the associated Laplacian or its determinant. 
On the other hand, the advantage of using  $\hat{s}_{\tmop{VN}}$ over $s_{\tmop{VN}}$ and $s_{ST}$ is its much faster $O(|E|]$ and possible incremental calculation depending only on the node degrees which we plan to investigate further as it would enable real-time performance of \tbsp. Also visible from Fig. \ref{fig:timePerAction} is that \tbsp error bounds calculation adds a small additional cost, which is especially true for Hadamard bound.
%
%
%
While \tbsp was action consistent in all sessions for all proposed topological metrics ($\epsilon(J, s) = 0$), action trend consistency ($\gamma = 0$) was kept only for $s_{ST}$ in all cases, while only in the first planning session for $s_{VN}$ and $\hat{s}_{VN}$. $\gamma \neq 0$ can happen when topologies among actions are very similar and then other factors, e.g. noise or geometry, determine the solution.  However, $\gamma$ was still small. As we already stated, the determined bounds of the entropy can always provide a globally optimal solution but at the cost of evaluating the objective function of actions inside them, i.e. the ones whose lower bound is below an upper bound of the selected action.
From Figs. \ref{fig:EntropyBoundsSession1_01}-\ref{fig:EntropyBoundsSession2_05} we can see that a negative topological metric $s_{ST}$ is a very good approximation of the estimated posterior entropy even with large orientation noise, i.e. large $\xi$ values. On the other hand, its determined lower bound is very sensitive to it. For larger $\xi$ values, bounds are less tight and therefore our performance guarantees are more conservative. In practice, however, more important is that the upper bound is close to the real entropy since we make decisions based on it. We can also notice that for higher $\xi$, Hadamard approximation of the lower bound is getting better, meaning that we practically have very small computational cost in determining the bounds.

\begin{figure}
	\centering
	\subfloat[speed ratio $\doteq t_{ST}/t_{VN}$]{\includegraphics[width= 0.5\columnwidth]{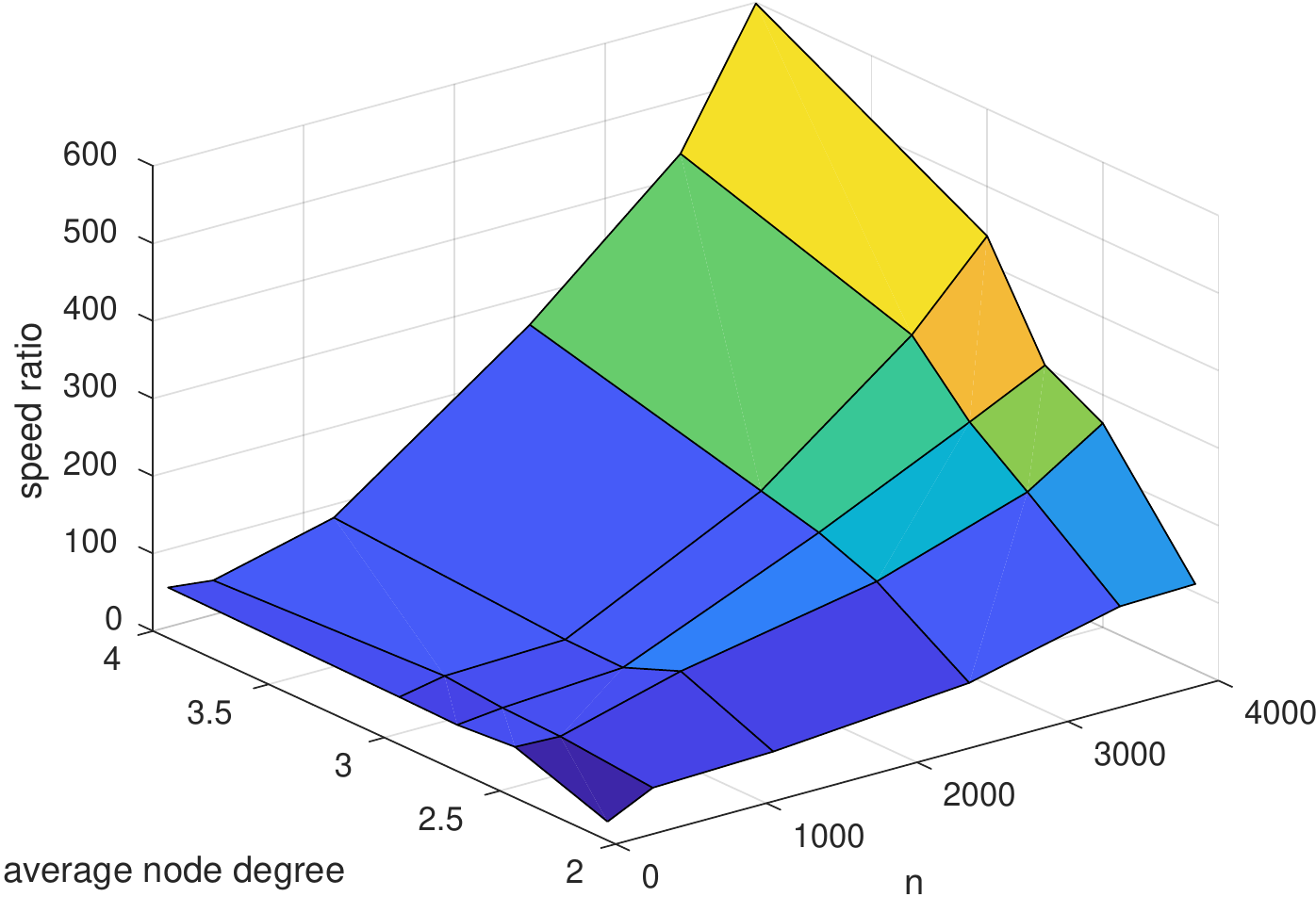}\label{fig:speed_simG}}
	\subfloat[graph signature vs. joint entropy of the FIM]{\includegraphics[width= 0.5\columnwidth]{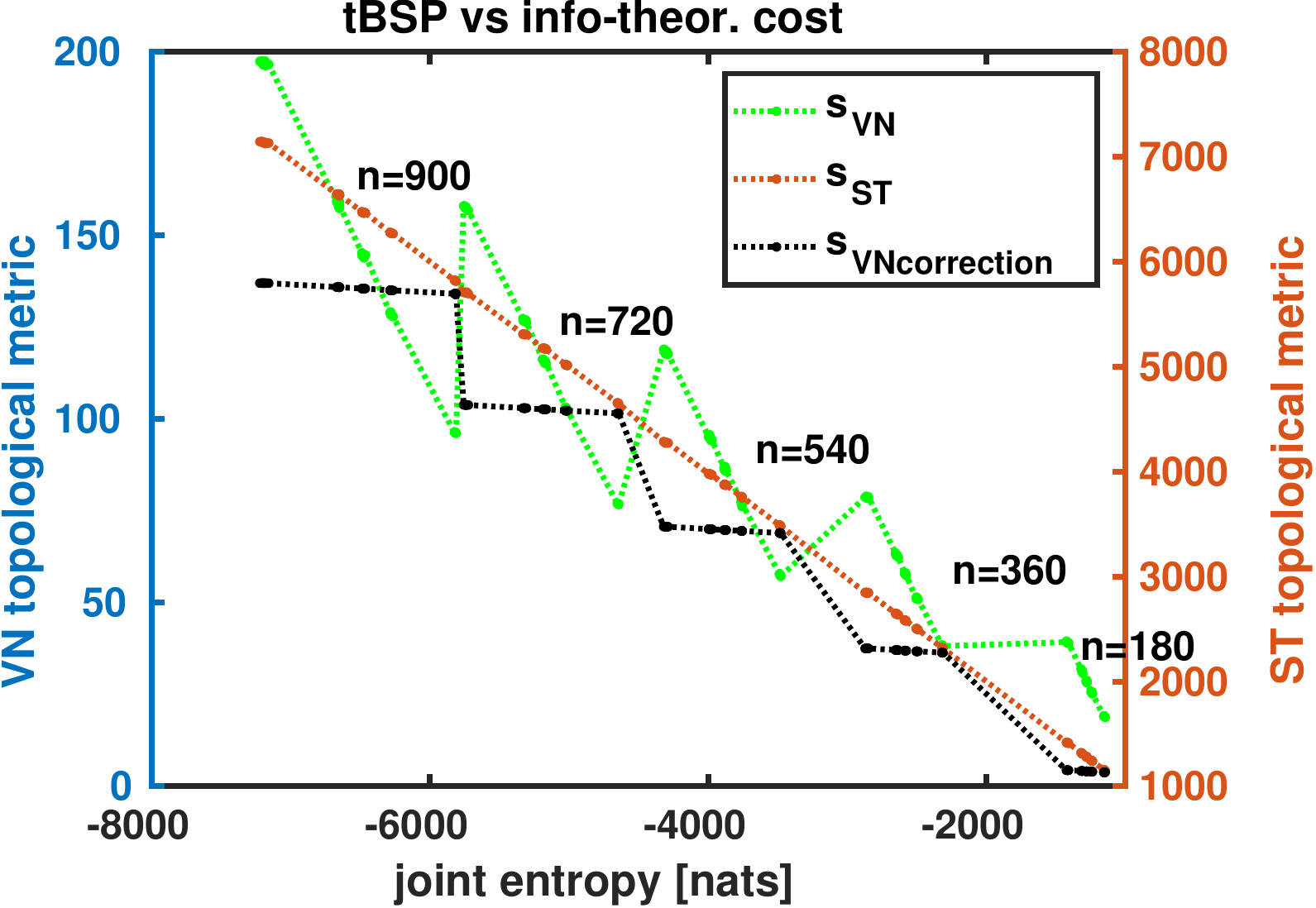}\label{fig:s_both_simG}}
	
	\caption{Performance on simulated random graphs.}
	\label{fig:SimRandomGraphs}
\end{figure}

In the next simulation, we show that good action consistency is still kept even when comparing significantly varying pose graphs.
We generated pose graphs with varying number of nodes $n$ and edges $m$ of which $n-1$ correspond to odometry and the rest to $LC = m-n+1$ loop closing factors. We choose  
$LC=\Delta \bar{d}\ n/2$, causing average node degree of a graph increasing in steps of $\Delta \bar{d} \in \left\lbrace 0.1, 0.5, 0.75, 1, 2\right\rbrace $ from a given odometry graph of size $n$ (whose average node degree is approx. 2).
For each pair $(n,m)$ we create a group of 10 random topologies, calculate their topological metrics $s_{ST}$ and $\hat{s}_{VN}$, and measure the average time for their calculation, $t_{ST}$ and $t_{VN}$ respectively. 
The results of this simulation are given in Fig. \ref{fig:SimRandomGraphs}.
We can see that correlation with the posterior entropy of FIM of MLE is still high for $s_{ST}$ on the whole range of topologies while $\hat{s}_{VN}$ follows the overall trend and is high for graphs with the same $n$, but breaks the action consistency with jumps of $n$ (Fig. \ref{fig:s_both_simG}). This suggests that normalization $f$ of $\hat{s}_{VN}$ should be applied according to its relation to the objective according to the Definition 1. For now, we only observe empirically that the normalization by adding the second term in eq. \eqref{eq:STsig} improves actions consistency. Although less accurate, the plot in Fig. \ref{fig:speed_simG} demonstrates why $\hat{s}_{VN}$ might still be interesting to use for fast decision making in some situations. Clearly, the gain in speed is significant in high dimensional BSP problems.

\section{CONCLUSIONS}

This paper provides theoretical foundations for action consistent topological belief space planning (\tbsp). \tbsp enables efficient decision making in high dimensional state spaces by considering only topologies of factor graphs that correspond to posterior future beliefs. As such, it uses a simplified representation of the information-theoretic cost that can lead to a loss of performance. This approach is in contrast to existing approximate BSP solutions that aim to simplify the beliefs in such a way that enables more efficient objective calculation. We derive error bounds of \tbsp which can be calculated online, i.e.~with a small additional cost to the topological metric. One can then resort to \tbsp to drastically reduce computational cost while carefully monitoring a conservative estimate on the sacrifice in performance, that would be provided by these bounds, or to guarantee global optimality of \tbsp by evaluating generally a much smaller number of candidate actions.
A topological metric, an approximation of the von Neumann entropy of a graph  suggests a possible online BSP performance, but its relation to information-theoretic cost requires further investigation.








\bibliographystyle{plain}
\bibliography{../../references/refs,newrefs}

\end{document}